\documentclass[10pt,journal,compsoc]{IEEEtran}
\usepackage{cite}
\ifCLASSINFOpdf
   \usepackage[pdftex]{graphicx}
\else
\fi
\usepackage{amsmath}
\usepackage{amssymb}
\usepackage{tabulary,xfrac,enumitem}
\usepackage[dvipsnames]{xcolor}
\usepackage[ruled,noresetcount]{algorithm2e}
\usepackage[pagebackref=false,breaklinks=true,colorlinks,bookmarks=false]{hyperref}
\definecolor{citecolor}{RGB}{34,139,34}

\usepackage{array}
\usepackage{url}

\usepackage{booktabs}
\usepackage{multirow}
\usepackage{threeparttable}
\usepackage{color, colortbl}
\usepackage{changes}
\usepackage{subcaption}
\usepackage{wrapfig}
\usepackage{diagbox}
\usepackage{arydshln} 
\usepackage{makecell}

\usepackage{pifont}

\makeatletter
\newcommand{\thickhline}{%
    \noalign {\ifnum 0=`}\fi \hrule height 0.5pt
    \futurelet \reserved@a \@xhline
}
\makeatother
\definecolor{LightGray}{gray}{0.9}

\newcommand{\pub}[1]{{\color{gray}{\tiny{[{#1}]}}}}

\newcommand{\textblue}[1]{\textcolor{black}{#1}}

\begin{document}
%
\title{MIGC++: Advanced Multi-Instance Generation Controller for Image Synthesis}
%
%
%
%

\newcommand{\zongxin}[1]{{#1}}
\newcommand{\new}[1]{{#1}}

\author{Dewei~Zhou,
        You Li,
        Fan Ma,
        Zongxin Yang,
        and~Yi~Yang
\thanks{

This work was supported by National Natural Science Foundation of China (U2336212) and Fundamental Research Funds for the Zhejiang Provincial Universities (226-2024-00208). (Corresponding author: Zongxin Yang.)

D. Zhou, Y. Li, F. Ma, and Y. Yang are with ReLER, CCAI, Zhejiang University, Hangzhou, 310027, China (e-mail: \{zdw1999, uli2000, mafan, yangzongxin, yangyics\}@zju.edu.cn).

Z. Yang is with DBMI, HMS, Harvard University, Boston 02115, USA. (e-mail: \{zdw1999, uli2000, mafan, yangzongxin, yangyics\}@zju.edu.cn).

Thank Ji Xie for extensive experiments, excellent demos in consistent-MIG.
}}

\markboth{IEEE TRANSACTIONS ON PATTERN ANALYSIS AND MACHINE INTELLIGENCE}%
{Shell \MakeLowercase{\textit{et al.}}: Bare Demo of IEEEtran.cls for Computer Society Journals}
\IEEEtitleabstractindextext{%
\begin{abstract}

%

%
%
%
%
%
%
We introduce the Multi-Instance Generation (MIG) task, which focuses on generating multiple instances within a single image, each accurately placed at predefined positions with attributes such as category, color, and shape, strictly following user specifications. MIG faces three main challenges: avoiding attribute leakage between instances, supporting diverse instance descriptions, and maintaining consistency in iterative generation. To address attribute leakage, we propose the Multi-Instance Generation Controller (MIGC). MIGC generates multiple instances through a divide-and-conquer strategy, breaking down multi-instance shading into single-instance tasks with singular attributes, later integrated. To provide more types of instance descriptions, we developed MIGC++. MIGC++ allows attribute control through text \& images and position control through boxes \& masks. Lastly, we introduced the Consistent-MIG algorithm to enhance the iterative MIG ability of MIGC and MIGC++. This algorithm ensures consistency in unmodified regions during the addition, deletion, or modification of instances, and preserves the identity of instances when their attributes are changed.  We introduce the COCO-MIG and Multimodal-MIG benchmarks to evaluate these methods. Extensive experiments on these benchmarks, along with the COCO-Position benchmark and DrawBench, demonstrate that our methods substantially outperform existing techniques, maintaining precise control over aspects including position, attribute, and quantity.
Project page: \url{https://github.com/limuloo/MIGC}.

\end{abstract}

}

\maketitle
\IEEEdisplaynontitleabstractindextext
%
\IEEEpeerreviewmaketitle

\IEEEraisesectionheading{\section{Introduction}\label{sec:introduction}}

Stable Diffusion (SD) possesses the capability to generate images based on textual descriptions and is currently widely used in fields such as gaming, painting, and photography\cite{diffusionrig,123,controlnet,caphuman,quan2024psychometry,zhang2023sifu,yang2024doraemongpt}. Recent research on SD has primarily concentrated on single-instance scenarios~\cite{dragdiffusion,dreambooth,huang2023avatarfusion,xu2023seeavatar,anydoor,selfdiffusion,instantbooth,pfbdiff,imagic,holodiff,123}, where the models render only one instance with a single attribute. However, when applied to scenarios requiring the simultaneous generation of multiple instances, SD faces challenges in offering precise control over aspects like the positions, attributes, and quantity of the instances generated~\cite{aae,structurediff,dividebind,conform,migc}.

\begin{figure}[t!]

\centering
    \includegraphics[width=1.0\linewidth]{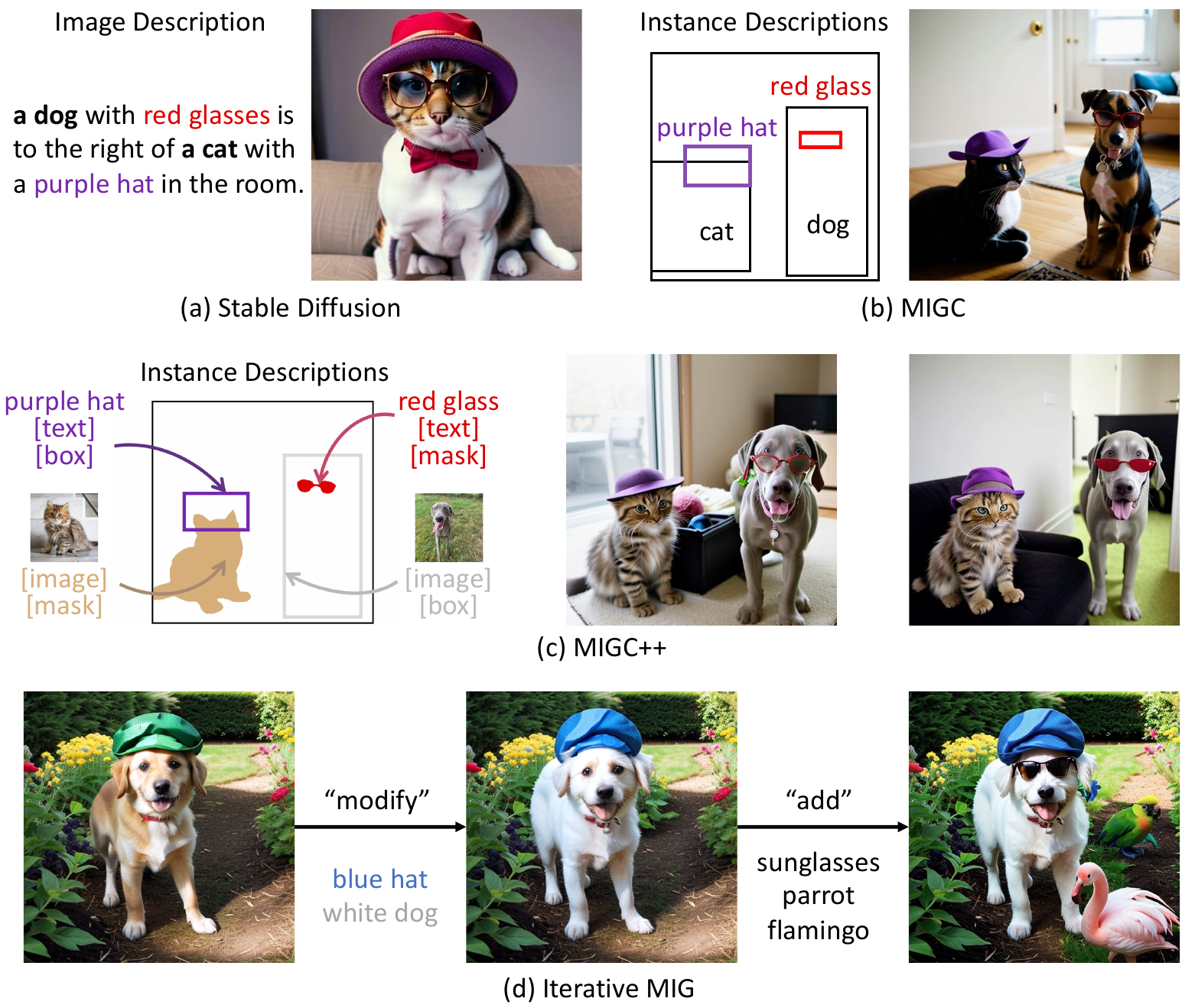}

\vspace{-2.5mm}
    \caption{
    \textbf{Illustration of MIG.} (a) SD generates images from a single image description, struggling with position control (e.g., locating a missing dog) and attribute control (e.g., incorrect hat color) in MIG. (b) MIGC ensures precise attribute and positional fidelity by using bounding boxes for spatial definitions and text for attribute definitions. (c) MIGC++ extends the framework's versatility, integrating both textual and visual descriptors for attributes and employing bounding boxes and masks to define positions. (d) Building on MIGC and MIGC++, we introduce the Consistent-MIG algorithm to bolster iterative MIG capabilities. }
    \label{fig:mig_overview}
    \vspace{-3.5mm}

\end{figure}

We introduce the Multi-Instance Generation (MIG) task, designed to advance the capabilities of generative models in multi-instance scenarios. MIG requires models to generate each instance according to specific attributes and positions detailed in instance descriptions, while maintaining consistency with the global image description. The MIG is illustrated in Fig.~\ref{fig:mig_overview}(b) and Fig.~\ref{fig:mig_overview}(c), offering distinct advantages over Text-to-Image methods such as SD~\cite{stablediffusion} in Fig.~\ref{fig:mig_overview}(a).

%
There are three main problems when applying SD-based methods to the MIG: \textbf{1) Attribute leakage}: CLIP~\cite{CLIP}, the text encoder of SD, allows inter-instance attribute interference~\cite{structurediff} (e.g., a ``purple hat'' influenced by ``red glasses'' in Fig.~\ref{fig:mig_overview}(a)). Also, SD's Cross-Attention~\cite{attention} lacks precise localization, causing misapplied \textit{shading}~\cite{migc} 
across instances (e.g., shading intended for a missing dog affects a cat's body in Fig.~\ref{fig:mig_overview}(a)). \textbf{2) Restricted instance description}: The current methods typically only allow for describing instances using a single modality, either text~\cite{instancediffusion,reco,gligen} or images~\cite{elite,ssr,multidiffusion}, which restricts the freedom of user creativity. Additionally, current methods~\cite{boxdiff,layoutdiff,reco,gligen} primarily use bounding boxes to specify instance locations, which limits the precision of instance position control. \textbf{3) Limited iterative MIG ability}: During the addition, deletion, or modification of instances in MIG, unmodified regions are prone to changes (see Fig.~\ref{fig:consistent_mig_ablation}(a)). Modifying instances' attributes, such as color, can inadvertently alter their ID (see Fig.~\ref{fig:consistent_MIG_vs_bld}(a)).

\begin{figure}[tb]
    \centering
    \includegraphics[width=1.0\linewidth]{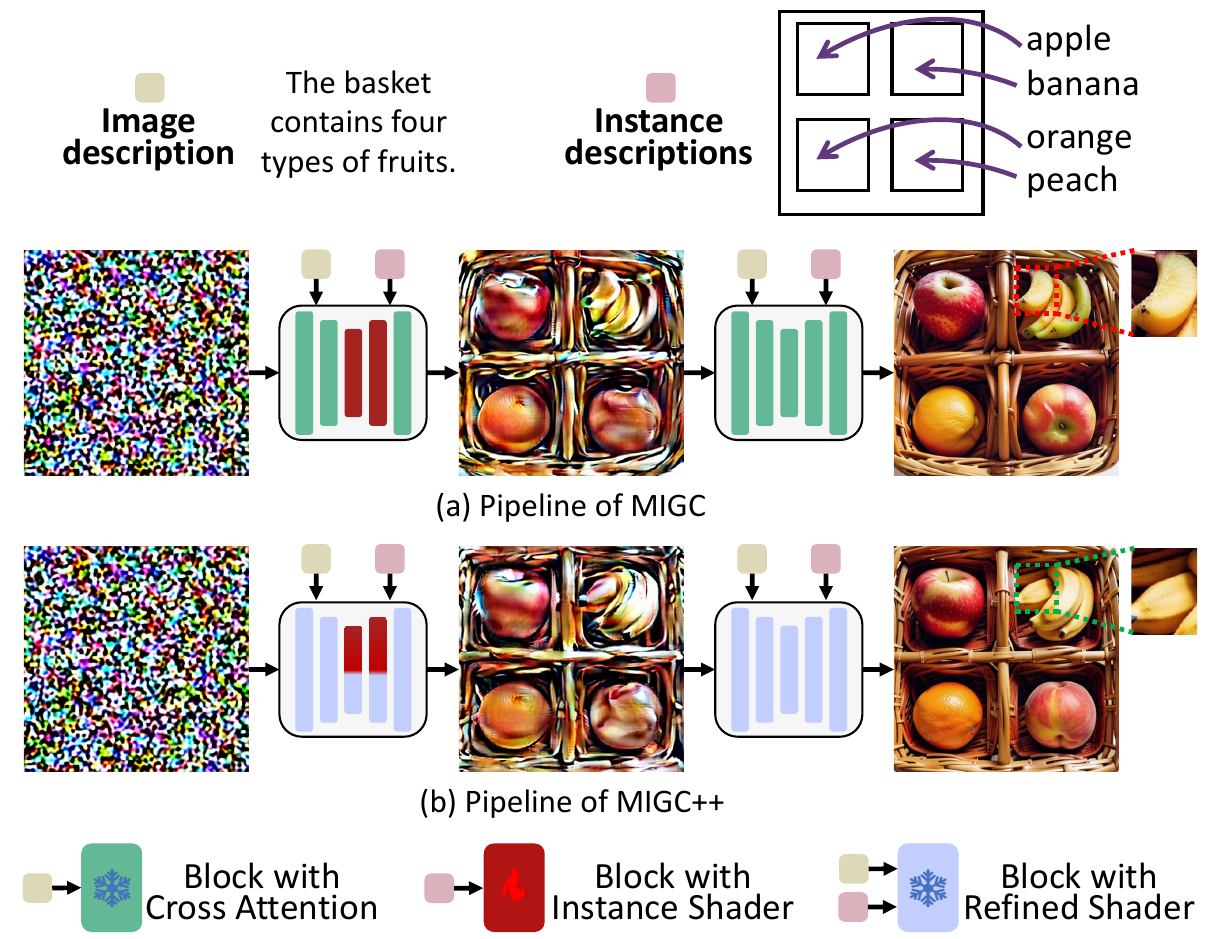}

\vspace{-1.5mm}
    \caption{
    \textbf{Comparison of the MIGC and MIGC++.} (a) MIGC incorporates Instance Shaders in the U-net's mid-block and deep up-blocks during high-noise sampling to ensure positional and coarse attribute control. (b) In addition to allowing more formats of describing instances (see Fig.~\ref{fig:mig_overview}(c)), MIGC++ introduces training-free Refined Shaders that supplant the Cross-Attention layers, enhancing accuracy in fine-grained details (e.g., better "banana" details).}
    \label{fig:migc_overview}
    
\vspace{-3.5mm}

\end{figure}

To \textbf{address attribute leakage}, we introduce the Multi-Instance Generation Controller (MIGC). Unlike existing methods~\cite{attention,gligen,instancediffusion,reco} that employ a single Cross-Attention mechanism for direct multi-instance shading, which can lead to attribute leakage, MIGC divides the task into separate single-instance subtasks with singular attributes and integrates their solutions.
Specifically, the MIGC employs Instance Shaders within the mid-block and deep up-blocks of the U-net architecture~\cite{Unet}, as shown in Fig.~\ref{fig:migc_overview}(a). 
Each shader begins with an Enhance-Attention mechanism for attribute-correct single-instance shading, followed by a Layout-Attention mechanism to create a shading template that bridges individual instances. Finally, a Shading Aggregation Controller combines these results into an attribute-correct multi-instance shading output, as shown in Fig.~\ref{fig:instanceshader}.

%

\begin{figure}[tb]
    \centering
    \includegraphics[width=1.0\linewidth]{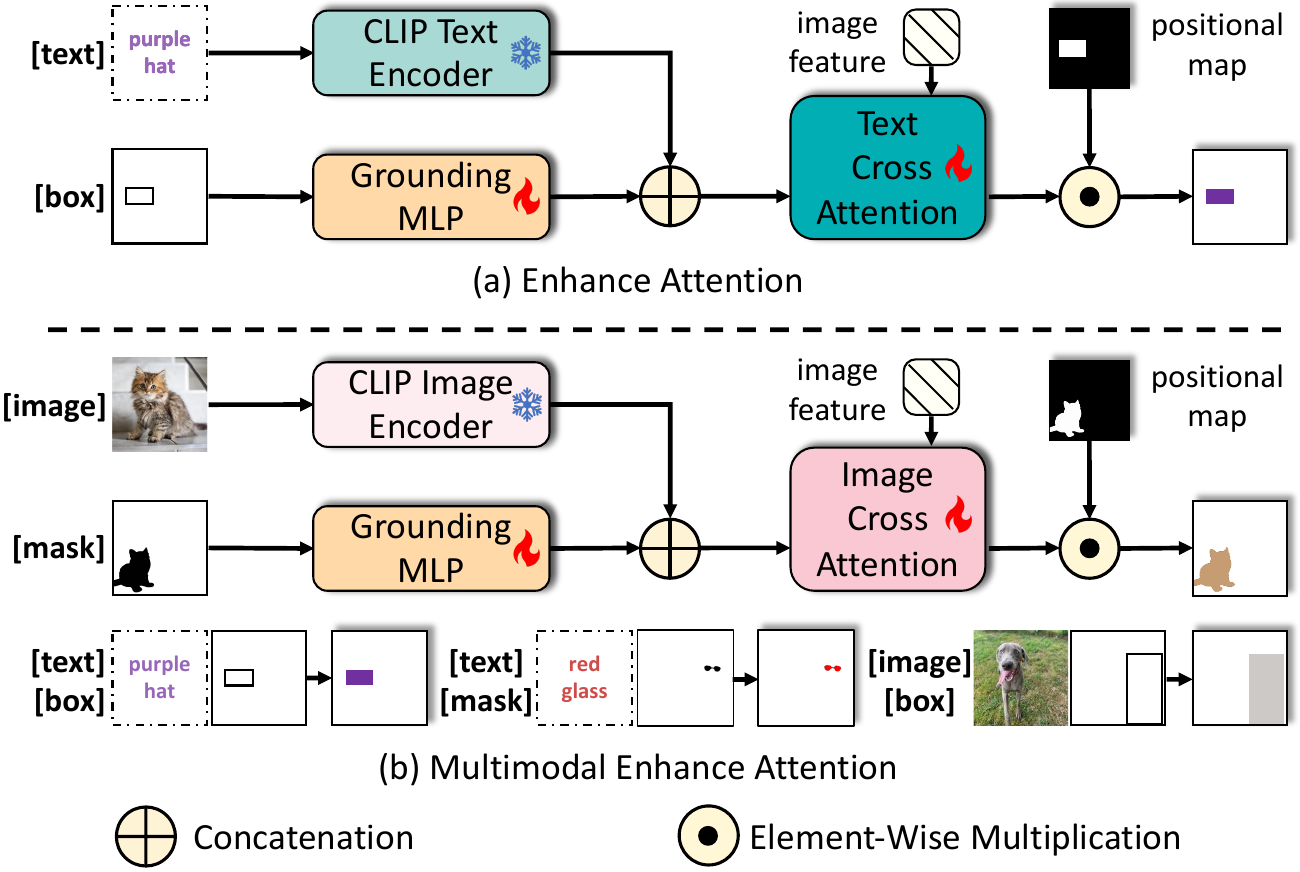}

\vspace{-1.5mm}
    \caption{\textbf{Enhance Attention (\S\ref{sec:ea}, \S\ref{sec:mmea})} enhances text embeddings to grounding embeddings, utilizing a trainable Cross-Attention layer for single-instance shading (a). The original MIGC approach can only describe an instance with text and bounding boxes. Building on this, MIGC++ expands this framework to Multimodal Enhanced Attention, enabling the description of an instance with various modalities within one generation (b).}
    \label{fig:enhanceattn}

\vspace{-3.5mm}
\end{figure}

To further \textbf{enrich the format of instance descriptions}, we have developed an advanced version of MIGC, termed MIGC++.
The MIGC++ extends the capabilities of MIGC by allowing attribute descriptions to transition from textual formats to more detailed reference images and localization from bounding boxes to finer-grained masks, as illustrated in Fig.~\ref{fig:mig_overview}(c). 
MIGC++ achieves this with a Multimodal Enhance-Attention (MEA) mechanism. As shown in Fig.~\ref{fig:enhanceattn}, by employing different weighted Enhance-Attention mechanisms, the MEA handles parallel shading for each instance across various modalities. By converting positional information into a unified 2D position map, it can accurately locate instances across multiple positional formats. Additionally, to enhance attribute detailing, crucial for image modal control, MIGC++ introduces a training-free Refined Shader for detailed shading, as depicted in Fig.~\ref{fig:migc_overview}(b) and Fig.~\ref{fig:refinedshader_effect}.

\begin{figure}[tb]
    \centering
    \includegraphics[width=1.0\linewidth]{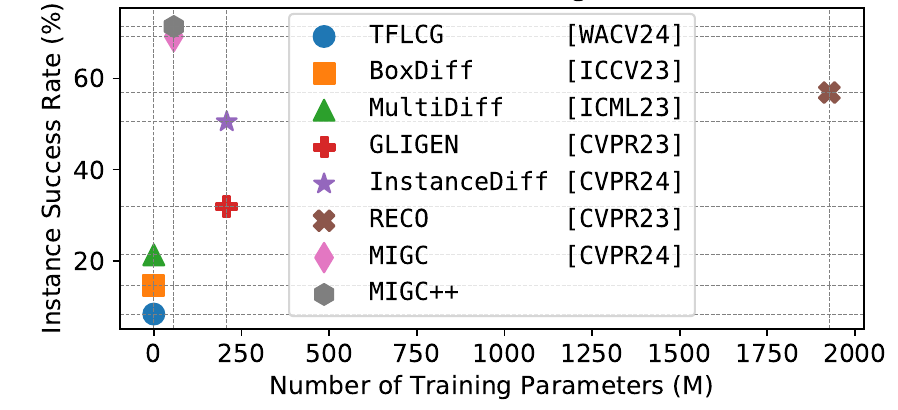}
    \vspace{-5.5mm}

    \caption{\textbf{Performance vs. Training Parameters.} Our MIGC and MIGC++ outperformed all competitors and required the fewest parameters among methods that need training.}
    \label{fig:performance}
    
\vspace{-3.5mm}

\end{figure}

To lastly \textbf{enhance the Iterative MIG ability} of MIGC and MIGC++, the Consistent-MIG algorithm is proposed.
This algorithm replaces unmodified areas with results from the previous iteration to maintain background consistency, as shown in Fig.~\ref{fig:mig_overview}(d) and Fig.~\ref{fig:consistent_mig_ablation}(b). During the Self-Attention~\cite{attention,stablediffusion} phase, it concatenates the prior iteration's Key and Value to preserve the ID consistency of the modified instance, as shown in Fig.~\ref{fig:mig_overview}(d) and Fig.~\ref{fig:consistent_MIG_vs_bld}(b).

To investigate the MIG task, we developed the COCO-MIG benchmark and an automated evaluation pipeline. This benchmark advances beyond traditional layout-to-image benchmarks like COCO~\cite{coco} by requiring simultaneous control over the positioning, attributes, and quantity of instances. COCO-MIG is divided into COCO-MIG-BOX and COCO-MIG-MASK, based on the format for indicating instance positions. Furthermore, we introduced the MultiModal-MIG benchmark to assess the capability of models in synthesizing multiple instances described by both textual and visual inputs.


We evaluated our approach using a suite of benchmarks, including our own COCO-MIG-BOX, COCO-MIG-MASK, and MultiModal-MIG, alongside established benchmarks such as COCO-Position~\cite{coco} and DrawBench~\cite{imagen}. On the COCO-MIG-BOX, MIGC improved the Instance Success Ratio (ISR) by \textbf{37.4}\% over GLIGEN~\cite{gligen}, with MIGC++ further enhancing this metric by an additional \textbf{2.2}\%, and as illustrated in Fig.~\ref{fig:performance}, MIGC and MIGC++ require fewest training parameters among methods that need training. In the COCO-Position benchmark, MIGC achieved a \textbf{14.01} increase in Average Precision over GLIGEN, with MIGC++ boosting this further by \textbf{8.87}. MIGC++ also led on COCO-MIG-MASK, surpassing Instance Diffusion~\cite{instancediffusion} by \textbf{16}\% in ISR. On DrawBench, MIGC++ substantially outperformed existing text-to-image methods, attaining an attribute control accuracy of \textbf{98.5}\%. In the MultiModal-MIG benchmark, MIGC++, as a tuning-free model, outstripped SSR-Encoder~\cite{ssr} by \textbf{74}\% in text-image alignment. This comprehensive testing demonstrates the robustness and versatility of our model across diverse generative tasks.

Our contributions in this paper are several folds:
\begin{itemize}[leftmargin=*, noitemsep, topsep=0pt]

\item We introduce the Multi-Instance Generation (MIG) task, which expands Single-Instance Generation to more complex and realistic applications in vision generation.

\item Utilizing a divide-and-conquer strategy, we propose the novel MIGC approach. This plug-and-play controller significantly enhances the MIG capabilities of SD models, providing precise control over the position, attributes, and quantity of instances in the generated image.

\item We developed the advanced MIGC++ approach, enabling simultaneous use of text and images to specify instance attributes, and employing boxes and masks for positioning. To our knowledge, MIGC++ is the first approach integrating these features.

\item We introduced the Consistent-MIG algorithm, which enhances the iterative MIG capabilities of MIGC and MIGC++, ensuring consistency across non-modified regions and the identity of modified instances.

\item We established the COCO-MIG benchmark to study the MIG task and formulated a comprehensive evaluation pipeline. We also launched the Multimodal-MIG benchmark to assess model capabilities in controlling instance attributes using both text and images simultaneously.

\begin{figure}[tb]
    \centering
    \includegraphics[width=1.0\linewidth]{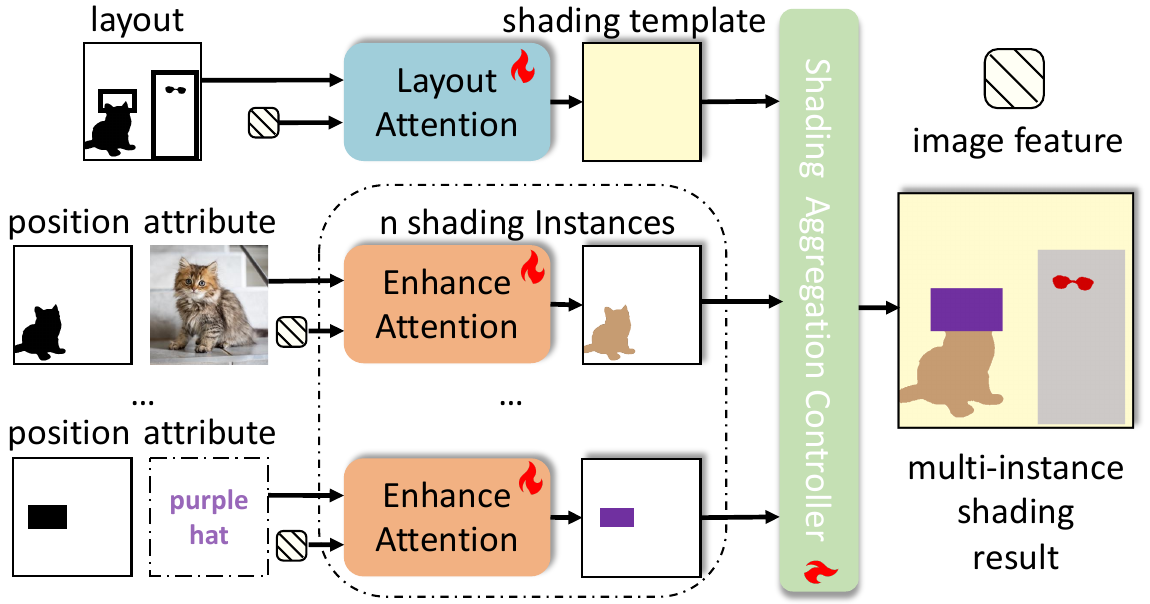}

\vspace{-1.5mm}
    \caption{\textbf{Overview of the proposed Instance Shader (\S\ref{sec:instance_shader}).} 
    The Instance Shader employs the Enhance-Attention mechanism for precise single-instnace shading, uses Layout-Attention to create a cohesive template, and concludes with a Shading Aggregation Controller for unified results.
    }
    \label{fig:instanceshader}
    
\vspace{-3.5mm}

\end{figure}

\end{itemize}

This paper expands upon our conference paper~\cite{migc}. 
Extensions are presented in various aspects. \textbf{In Method: (1)} We provided a more detailed explanation of MIGC (\S\ref{sec:migc}) and further elaborated on its deployment (\S\ref{sec:migc_deployment}). \textbf{(2)} We introduced a stronger version of MIGC, named MIGC++ (\S\ref{sec:migc++}), which allows users to describe instance attributes using both text and images, and instance positions using both masks and boxes. MIGC++ introduced a Refined Shader (\S\ref{sec:refined_shader}) to better control detailed attributes. \textbf{(3)} We proposed a Consistent-MIG (\S\ref{sec:consistent_mig}) algorithm to enhance the iterative MIG capability of MIGC and MIGC++. \textbf{In Benchmark: (4)} We refined the COCO-MIG benchmark (\S\ref{sec:cocomig}), splitting it into COCO-MIG-BOX and COCO-MIG-MASK. \textbf{(5)} We proposed a Multimodal-MIG (\S\ref{sec:multimodalmig_bench}) benchmark to assess the control over instance attributes using both text and image modality. \textbf{In Experiment: (6)} We compared with the current SOTA methods on COCO-MIG-BOX and COCO-MIG-MASK (\S\ref{sec:compare}), adding more robust baseline~\cite{instancediffusion,reco}. \textbf{(7)} We conducted experiments on Multimodal-MIG (\S\ref{sec:compare}). \textbf{(8)} We conducted more ablation experiments (\S\ref{sec:ablation}), including the deployment schemes of MIGC and MIGC++, the effectiveness of Refined Shader, the effectiveness of Consistent-MIG.


\section{Related Work}\label{sec:related_works}

\noindent\textbf{Diffusion models}~\cite{ddpm,pydiff} generate high-quality images by iteratively denoising Gaussian noise, but this process initially requires numerous iterations, slowing down image generation. To expedite this process, several training-free samplers—such as the DDIM~\cite{ddim}, Euler~\cite{elud}, and DPM-Solver~\cite{dpm} samplers—have been developed. Compared to the original DDPM~\cite{ddpm}, which necessitated more iterations, these samplers enable more efficient image generation with fewer steps and improved image quality. Additionally, Stable Diffusion~\cite{stablediffusion} techniques have been advanced to perform the denoising process in a compressed VAE~\cite{vqvae2,videoGPT,autoencode,taming} latent space, enhancing both training and sampling speeds.

\noindent\textbf{Text-to-image} generation aims to create images from textual descriptions. Initially, conditional Generative Adversarial Networks (GANs)~\cite{gan,attngan,crossmodal,zc_cycle} were used for this purpose. However, diffusion models~\cite{glide,stablediffusion,imagen,reimagen,eDiff-I,DALL-E2,Ho2022ClassifierFreeDG,pydiff,zhao2023wavelet,zhao2024learning,lu2023tf,lu2024mace,texface} and autoregressive models~\cite{MuseTG,cogview,parti} have largely replaced GANs, offering more stable training and enhanced image quality. To control generated content, Guided Diffusion~\cite{guideddiffusion} utilizes classifiers to assess generated images and guidances the sampling through the gradient of the classifiers. Classifier-free guidance (CFG)~\cite{Ho2022ClassifierFreeDG} further optimizes this process by interpolating between conditioned and unconditioned predictions, with methods like GLIDE~\cite{glide}, showing promising results under these protocols. DALL-E 2~\cite{DALL-E2} employs transformations from CLIP text features~\cite{CLIP} to CLIP image space for image generation, while Imagen~\cite{imagen} leverages the T5 large language model~\cite{T5} as a text encoder to improve quality. 
eDiff-I~\cite{eDiff-I} further elevates image quality by integrating expert generators. 
Stable Diffusion~\cite{stablediffusion} incorporates the attention mechanism to seamlessly integrate textual information into the generation process.

\noindent\textbf{Layout-to-image} generation techniques refine the positional accuracy of Text-to-image methods by integrating layout information. GLIGEN~\cite{gligen} and InstanceDiffusion~\cite{instancediffusion} advance this approach by expanding text tokens to include positional data as grounded tokens, which are further integrated into image features via an additional gated self-attention layer. Building on this foundation, LayoutLLM-T2I~\cite{layoutllm} enhances the GLIGEN framework with a relation-aware attention module, while RECO~\cite{reco} effectively combines layout and textual information to refine spatial control in generation processes. Additionally, certain models~\cite{tflcg,boxdiff,layoutdiff} have achieved training-free layout control within large-scale text-to-image systems by utilizing cross-attention maps to calculate layout loss, which directs the image generation process in a classifier-guided manner. Despite these technological improvements, managing the attributes of instances in MIG remains a challenge, often leading to images with blended attributes. This paper presents the MIGC and MIGC++ methods, developed to meticulously control the position, attribute, and quantity of generated instances in MIG.


\section{Preliminary: Stable Diffusion}\label{sec:revisit}

\noindent\textbf{CLIP text encoder.} 
SD~\cite{stablediffusion} uses the CLIP text encoder~\cite{CLIP} to encode the given image description $\boldsymbol{c}$ as a sequence of text embeddings $\textbf{W}_{c}=\mathrm{CLIP_{text}}(\boldsymbol{c})$. However, the contextualization of CLIP text embedding may cause attribute leakage between instances in the multi-instance scenarios~\cite{structurediff}.

\noindent\textbf{Cross-Attention layers.}
SD uses the Cross-Attention~\cite{attention} mechanism to inject text information into the 2D image feature $\mathbf{X} \in \mathbb{R}^{(H, W, C)}$. By using a linear layer $f^{Q}_{c}(\cdot)$ to project image features as a query $\mathbf{Q}_{c}$, and using two linear layers $f^{K}_{c}(\cdot), f^{V}_{c}(\cdot)$ to project text embeddings $\textbf{W}_{c}$ into key $\mathbf{K}_{c}$ and value $\mathbf{V}_{c}$,  the Cross Attention can be formulated as:
\begin{equation}\small
\label{eq:cross_text}
\mathbf{R}_{c}=\mathrm{\mathbf{CA}}(\mathbf{X},\boldsymbol{c})=\mathrm{\mathbf{Softmax}}\left(\frac{\mathbf{Q}_{c}\mathbf{K}_{c}^T}{\sqrt{d}}\right) \mathbf{V}_{c} ,
\end{equation}
where $\mathbf{R}_{c} \in \mathbb{R}^{(H, W, C)}$ is the output. The output $\mathbf{R}_{c}$ will be added to the original image feature $\mathbf{X}$ to determine the final content in generated images, and this process is like a shading~\cite{migc} operation in the image feature.

\noindent\textbf{Unet denoising network.} 
SD employs a U-net~\cite{Unet} architecture to predict the noise in the reverse process of the diffusion models~\cite{ddpm}. The U-net is segmented into down-blocks, one mid-block, and up-blocks~\cite{resnet,attention}, with Cross-Attention layers embedded within them. The down-blocks gradually reduce the resolution of the input image. The mid-block is at the deepest part of the U-net and is the main place to adjust the image content and layout. Up-blocks gradually restore the image resolution and complete image generation. The deeper image features of the mid-block and up-blocks contain rich semantic and layout information.

\noindent\textbf{Image Projector.} 
It's very hard to describe a customized concept only using a textual description. Approaches such as ELITE~\cite{elite}, IP-Adapter~\cite{ip-adapter}, and SSR-Encoder~\cite{ssr} employ a learning-based projector to integrate visual concepts into the image feature $\mathbf{X}$. Given a reference image $\boldsymbol{e}$, these methods utilize the CLIP image encoder~\cite{CLIP} to encode it as a sequence of image embeddings $\textbf{W}_{e}=\mathbf{\mathrm{CLIP}}_{image}(\boldsymbol{e})$. Then, a trainable projection network refines these embeddings to extract the fine-grained subject feature $\textbf{W}'_{e}=\mathbf{\mathrm{Proj}}(\textbf{W}_{e})$. Finally, the shading result $\mathbf{R}_{e}$ can be formulated as follows:

\begin{equation}\small
\label{eq:cross_image} \mathbf{R}_{e}=\mathrm{\mathbf{IP}}(\mathbf{X}, \boldsymbol{e})   =\mathrm{\mathbf{Softmax}}\left(\frac{\mathbf{Q}_c {\mathbf{K}_e}^T}{\sqrt{d}}\right) \mathbf{V}_{e},
\end{equation}
where key ${\mathbf{K}_e}$ and value ${\mathbf{V}_e}$ are derived by projecting the feature $\textbf{W}'_{e}$ through newly added linear layers $f_{e}^K(\cdot), f_{e}^V(\cdot)$, and the query ${\mathbf{Q}_{c}}$ remains the same as that in Eq.~\ref{eq:cross_text}.


\section{Methodology}\label{sec:method}

In this section, we introduce the Multi-Instance Generation (MIG), define the task, and outline its primary challenges. We then describe MIGC, which employs a divide-and-conquer strategy to accurately render instance layouts and coarse attributes, significantly mitigating attribute leakage. Building on this, we present MIGC++, which supports more flexible descriptions for individual instances in MIG and integrates a Refined Shader to enhance detailed shading. To further improve the iterative capabilities of MIGC and MIGC++, we propose the Consistent-MIG algorithm.

\vspace{-3.0mm}
\subsection{Multi-Instance Generation Task}\label{sec:mig}

\noindent\textbf{Definition.}
The Multi-Instance Generation (MIG) task not only provides a global description $\boldsymbol{c}$ for the target image but also includes detailed descriptions for each instance $\boldsymbol{i}$, specifying both its position ${pos}_{i}$ and attribute ${attr}_{i}$.
Generative models must ensure that each instance conforms to its designated position ${pos}_{i}$ and attributes ${attr}_{i}$, in harmony with the global image description $\boldsymbol{c}$.

\noindent\textbf{Problem.}\label{sec:mig_Challenge}
Current methodologies face three main problems in the MIG. $\textit{(i) Attribute Leakage.}$ The sequential encoding of the image description $\boldsymbol{c}$ with multiple attributes causes later embeddings to be influenced by earlier ones~\cite{CLIP,structurediff}. For example, as shown in Fig.~\ref{fig:mig_overview}(a), the purple hat appears red due to the influence of the preceding red glasses description. Additionally, the use of a single Cross-Attention operation~\cite{attention,stablediffusion} for multi-instance shading~\cite{migc} results in shading inaccuracies, such as the dog being inadvertently shaded onto the cat’s body, as illustrated in Fig.~\ref{fig:mig_overview}(a).
$\textit{(ii) Restricted instance description.}$ The current methods typically only allow for describing instances using a single modality, either text~\cite{instancediffusion,reco,gligen} or images~\cite{elite,ssr,multidiffusion}, which restricts the freedom of user creativity. Additionally, current methods~\cite{boxdiff,layoutdiff,reco,gligen} primarily use bounding boxes to specify instance locations, which limits the precision of instance position control. $\textit{(iii) Limited iterative MIG ability.}$ When modifying (e.g., adding or deleting) certain instances in MIG, unmodified regions are prone to changes (see Fig.~\ref{fig:consistent_MIG_vs_bld}.(a)). Modifying instances' attributes, such as color, can inadvertently alter their ID (see Fig.~\ref{fig:consistent_mig_ablation}(a)).

\noindent\textbf{Solution.} We initially introduce the MIGC \textbf{(\S~\ref{sec:migc})} method to address the issue of attribute leakage. Subsequently, we propose its enhanced version, MIGC++ \textbf{(\S~\ref{sec:migc++})}, which expands the forms of instance descriptions. Finally, we present Consistent-MIG \textbf{(\S~\ref{sec:consistent_mig})} to augment the iterative MIG capabilities of both MIGC and MIGC++.

\vspace{-3.0mm}
\subsection{MIGC}\label{sec:migc}
\subsubsection{Overview}\label{sec:instance_shader}
The MIGC method addresses attribute leakage by applying a divide-and-conquer strategy, which breaks down the complex multi-instance shading process into simpler, single-instance tasks. These tasks are processed independently in parallel to prevent attribute leakage. The outputs are then seamlessly merged, resulting in a coherent multi-instance shading result that is free from attribute leakage.

\noindent\textbf{Instance Shader} is designed according to the above divide-and-conquer approach, as depicted in Fig.~\ref{fig:instanceshader}. Initially, it divides multi-instance shading into multiple distinct single-instance shading tasks. Subsequently, an Enhance-Attention mechanism (\S\ref{sec:ea}) is applied to address each task individually, producing multiple shading instances. Following this, a Layout-Attention mechanism (\S\ref{sec:la}) is implemented to devise a shading template that facilitates the integration of individual instances. Ultimately, a Shading Aggregation Controller (\S\ref{sec:sac}) combines these instances and the template to produce the comprehensive multi-instance shading result. As depicted in Fig.~\ref{fig:migc_overview}, the MIGC replaces the Cross-Attention~\cite{attention,stablediffusion} layers with the Instance Shaders in both the mid-block and deep up-blocks of the U-Net~\cite{Unet} architecture to perform multi-instance shading on image feature, during high-noise-level sample steps, to enhance the accuracy of Multi-Instance Generation (\S\ref{sec:migc_deployment}).

\vspace{-2.0mm}
\subsubsection{Enhance Attention}\label{sec:ea}

{\textbf{Motivation.}} To achieve the single-instance shading, a straightforward method might involve leveraging the pre-trained Cross-Attention layers~\cite{attention,stablediffusion} in SD. However, this method encounters two significant problems, i.e., \textit{(i) Instance Merging}: Eq.~\ref{eq:cross_text} illustrates that when two instances share the same attributes, they own identical keys $\textbf{K}$ and values $\textbf{V}$ during the shading. If these instances are closely positioned or overlap, the latter combination may erroneously merge them into a single instance (see Fig.~\ref{fig:ablation_vis}(e)). \textit{(ii) Instance Missing}: Initial editing~\cite{initialimageedit} methods indicate that the initial noise largely influences the image layout in the SD's outputs. If the initial noise does not support an instance at the specified position, its shading result will be weak, leading to an instance missing (see Fig.~\ref{fig:ablation_vis}(a)). As shown in Fig.~\ref{fig:instanceshader}, Enhance Attention (EA) is devised to achieve accurate single-instance shading, solving the above two problems. 

\noindent{\textbf{Solution.}} To solve the \textit{instance merging} problem, the EA augments the attribute embeddings of each instance with position embeddings to identify instances with the same attribute but different positions.  As depicted in Fig.~\ref{fig:enhanceattn}(a), for an instance $\boldsymbol{i}$ described by ${text}_{i}$ and ${box}_{i}$, the EA first uses the CLIP text encoder~\cite{CLIP} to encode the textual attribute description ${text}_{i}$  as a sequence of text embedding $\textbf{W}^{i}_{text}$. Subsequently, the EA encodes the positional description ${box}_{i}=[x^{i}_{1},y^{i}_{1},x^{i}_{2},y^{i}_{2}]$ as position embedding $\textbf{W}^{i}_{pos}$ using a Grounding MLP, which incorporates a Fourier embedding transform~\cite{nerf} and an MLP layer: 
\setlength{\abovedisplayshortskip}{5pt}
\begin{equation}\small
\label{eq:pos_embed}
\textbf{W}^{i}_{pos}=\mathrm{\textbf{GroudingMLP}}(\boldsymbol{box}_{i})=\mathrm{\textbf{MLP}}(\mathrm{\textbf{Fourier}}(\boldsymbol{box}_{i})).
\vspace{-0.5mm}
\end{equation}
Finally, EA integrates attribute embedding and position embedding to form grounding embedding:

\setlength{\abovedisplayshortskip}{2pt}
\begin{equation}\small
\label{eq:grouding_embed}
\mathbf{G}_{i}=[\textbf{W}^{i}_{pos},\textbf{W}^{i}_{text}],
\vspace{-0.5mm}
\end{equation}
where $[\cdot,\cdot]$ represents the concatenation operation.

To solve the \textit{instance missing} problem, the EA uses a new trainable Cross Attention to perform enhancing shading, and the shading result can be formulated as: 

\begin{equation}\small
\label{eq:ea}
\mathbf{R}^{i}_{ea}=\mathrm{\mathbf{Softmax}}\left(\frac{\mathbf{Q}^{^{i}}_{ea}{\mathbf{K}^{i}_{ea}}^T}{\sqrt{d}}\right) \mathbf{V}^{i}_{ea} \cdot \mathbf{M}_{i} , \mathbf{R}^{i}_{ea} \in \mathbb{R}^{(H,W,C)},
\vspace{-0.5mm}
\end{equation}
where a learnable linear layer $f^{Q}_{ea}(\cdot)$ projects the image feature $\mathbf{X}$ as the query, two learnable linear layers $f^{K}_{ea}(\cdot), f^{V}_{ea}(\cdot)$ project the grounding embedding $\mathbf{G}_{i}$ as key and value, and the positional map $\mathbf{M}_{i}$ can be generated according to the $\boldsymbol{box}_{i}$, where values within the box region are set to 1, and all other values are set to 0. During the training phase, these positional maps facilitate precise spatial localization. This precise localization ensures that the shading effects produced by the EA are confined accurately to the targeted areas, which enables the EA to consistently apply shading enhancements across varying image features and effectively resolves the instance missing problem.

\subsubsection{Layout Attention}\label{sec:la}
\begin{figure}[tb]
    \centering
    \includegraphics[width=1.0\linewidth]{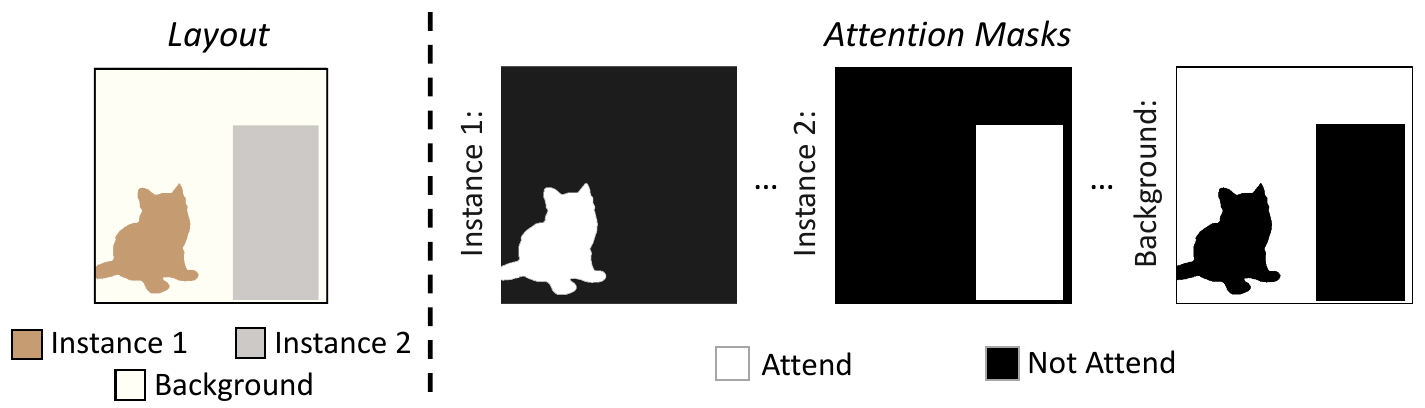}

\vspace{-1.5mm}
    \caption{\textbf{Layout Attention (\S\ref{sec:la})} operates akin to a Self-Attention mechanism but incorporates a layout constraint. This restriction ensures that each image token only attends to others located within the same instance region.}
    \label{fig:layoutattn}
    
\vspace{-3.5mm}

\end{figure}
\noindent{\textbf{Motivation.}} Utilizing the Enhance Attention to execute single-instance shading on the image features, we yield multiple individual instances. However, before merging them into the multi-instance shading result, a shading template is essential to bridge them, as the shading process of each instance is independent. As illustrated in Fig.~\ref{fig:instanceshader}, the Instance Shader employs Layout Attention to generate the shading template, which is conditioned on the layout of all instances.

\noindent{\textbf{Solution.}} The proposed Layout-Attention mechanism operates similarly to a Self-Attention~\cite{attention} process on image features $\mathbf{X}$, generating the shading template. It incorporates layout information to ensure that each image token interacts only with others within the same instance region, thereby mitigating attribute leakage across different instances, as depicted in Fig.~\ref{fig:layoutattn}. Given the layout information $\mathbb{M}=\left\{ \mathbf{M}_1, \ldots, \mathbf{M}_N, \mathbf{M}_{bg}\right\}$, where the instance mask $\mathbf{M}_{i}$ is defined in Eq.~\ref{eq:ea} and the background mask $\mathbf{M}_{bg}$ serves as the complementary mask to the instance masks, Layout Attention constructs the Attention Mask as:
%
\begin{equation}\small
\mathbf{A}_{la}^{p,q} = \begin{cases}
1,  \ \  \ \ \ \ \ \ \text{if}\, \exists \ \mathbf{M} \in \mathbb{M},\mathbf{M}_{p}= \mathbf{M}_{q}=1 \\
-inf,  \ \  \text{otherwise}
\end{cases},
  \label{eq:la_mask}
\end{equation}
where the value $\mathbf{A}_{la}^{p,q}$ determines whether the image token $\boldsymbol{p}$ attends to the image token $\boldsymbol{q}$ in the Attention operation, and the Layout Attention can be formulated as follows: 

\vspace{-3.5mm}

\begin{equation}
\small
  \mathbf{R}_{la} = \mathrm{\mathbf{Softmax}}(\frac{\mathbf{Q}_{la} {\mathbf{K}_{la}}^T}{\sqrt{d}} \odot \mathbf{A}_{la})\mathbf{V}_{la}, \mathbf{R}_{la} \in{\mathbb{R}^{(H,W,C)}},
  \label{eq:la}
\end{equation}
where $\odot$ represents the Hadamard product, and learnable linear layers $f^{K}_{la}(\cdot), f^{Q}_{la}(\cdot), f^{V}_{la}(\cdot)$ project the image feature $\mathbf{X}$ as key ${\mathbf{K}_{la}}$, query ${\mathbf{Q}_{la}}$, and value ${\mathbf{V}_{la}}$.

\begin{figure}[tb]
    \centering
    \includegraphics[width=1.0\linewidth]{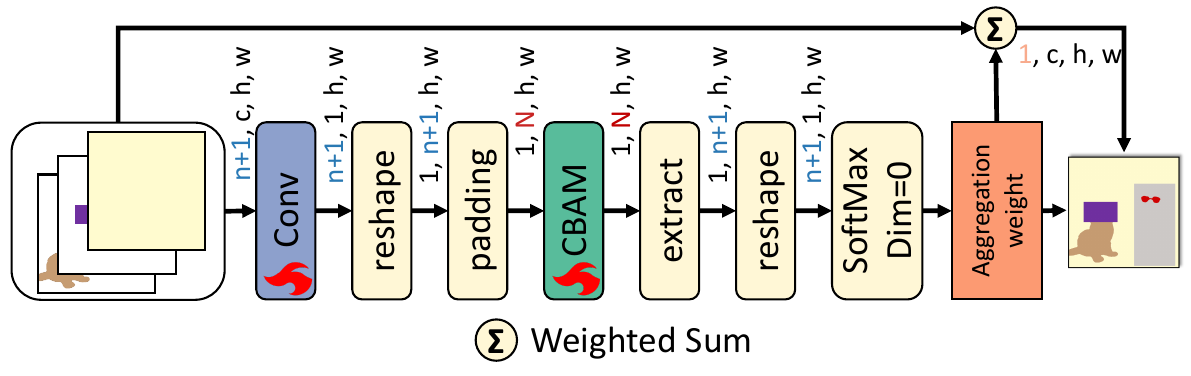}

    \vspace{-1.5mm}

    \caption{
    \textbf{Shading Aggregation Controller (\S\ref{sec:sac})} dynamically adjusts the aggregation weights of the n shading instances and the one shading template (i.e., total n+1 feature maps), ultimately producing the multi-instance shading result (i.e., one feature map).
    }
\vspace{-3.5mm}
    \label{fig:sac}

\end{figure}

\subsubsection{Shading Aggregation Controller}\label{sec:sac}
\noindent{\textbf{Motivation.}} To generate the final multi-instance shading result, we combine shading instances obtained via Enhance Attention with shading templates from Layout Attention. Dynamic adjustment of aggregation weights across blocks and sampling steps is crucial. Thus, we propose a Shading Aggregation Controller (SAC) to manage this process.

\noindent{\textbf{Solution.}} Fig.~\ref{fig:sac} presents the SAC framework. We begin by concatenating n shading instances ${\mathbf{R}^{1}_{ea}, \cdots, \mathbf{R}^{n}_{ea}}$ and a shading template $\mathbf{R}_{la}$ as inputs. The SAC uses a 1x1 convolution layer to extract initial spatial features from each instance. It then rearranges the feature dimensions and applies the Convolutional Block Attention Module (CBAM)~\cite{cbam} for instance-wise attention. Aggregation weights, normalized per spatial pixel, determine each instance's shading intensity, leading to the final result. To accommodate a variable number of shading instances, we set a predefined channel count, 
N, for CBAM, which is higher than the typical number of instances. During inference, features are zero-padded to this channel count, ensuring consistent processing by CBAM. This method enables dynamic adaptation to the actual number of shading instances, with SAC facilitating the multi-instance shading outcome:
\begin{equation}
\small
  \mathbf{R}_{inst} = \mathrm{\mathbf{SAC}}(\mathbf{R}^{1}_{ea}, \cdots, \mathbf{R}^{n}_{ea},\mathbf{R}_{la}), \mathbf{R}_{inst} \in{\mathbb{R}^{(H,W,C)}}.
  \label{eq:sac}
\end{equation}

\subsubsection{Deployment of MIGC}\label{sec:migc_deployment}
Replacing the original Cross-Attention~\cite{attention,stablediffusion} layers in the U-net with the proposed Instance Shader (\S\ref{sec:instance_shader}), MIGC is deployed at the mid-block and deep up-blocks for high noise-level sampling steps. The remaining Cross-Attention layer performs global shading based on the image description, as shown in Fig.~\ref{fig:migc_overview}(a). This deployment offers key advantages: 1) Reduced training costs and faster training due to fewer parameters. 2) Enhanced performance by focusing shading on critical image features, improving semantic integrity. The efficacy of MIGC's deployment strategy is also verified through an ablation study (see Tab.~~\ref{tab:instanceshader_pos_ablation}).


\vspace{-3.0mm}
\subsection{MIGC++}\label{sec:migc++}

\subsubsection{Overview}\label{sec:migc++_overview}
Our advanced MIGC++ allows users to describe instances using more diverse forms. It enables the specification of instance attributes through both text and image, and position definition using both box and mask, leveraging a Multimodal Enhance-Attention Mechanism \textbf{(\S~\ref{sec:mmea})}. Additionally, MIGC++ introduces a Refined Shader \textbf{(\S~\ref{sec:refined_shader})} for detailed shading, which is crucial when shading an instance according to a reference image.

\vspace{-2.0mm}
\subsubsection{Multimodal Enhance Attention}\label{sec:mmea}
In our framework, Enhance-Attention shades each instance in parallel (Fig.~\ref{fig:instanceshader}), facilitating different shading modalities within the same image. MIGC++ develops this into a Multimodal Enhance-Attention mechanism, employing diverse positional and attribute descriptions to manage multiple instances simultaneously (Fig.~\ref{fig:enhanceattn}(b)).

\noindent\textbf{For various position descriptions}, including bounding boxes and masks, the MEA first produces a corresponding 2D position map $\mathbf{M}_{i}$, in which pixels within the instance region are designated as 1, while all others are set to 0. This 2D position map is used to enable precise control over the shading region, ensuring targeted and accurate enhancement, as depicted in Eq.~\ref{eq:ea}. Then, to derive the position embedding, the MEA standardizes all position formats to a bounding box format and utilizes the GroundingMLP, as described in Eq.~\ref{eq:pos_embed}, to obtain the position embedding.

\noindent\textbf{For various attribute descriptions}, including text and images, the MEA utilizes separate Cross-Attention mechanisms to optimize shading. 
As demonstrated in Fig.~\ref{fig:enhanceattn}(a) and Fig.~\ref{fig:enhanceattn}(b), rather than using a single layer for both modalities, MEA applies tailored Cross-Attention layer to each modality, enhancing the outcome in each modality.

%

\begin{figure}[tb]
    \centering
    \includegraphics[width=1.0\linewidth]{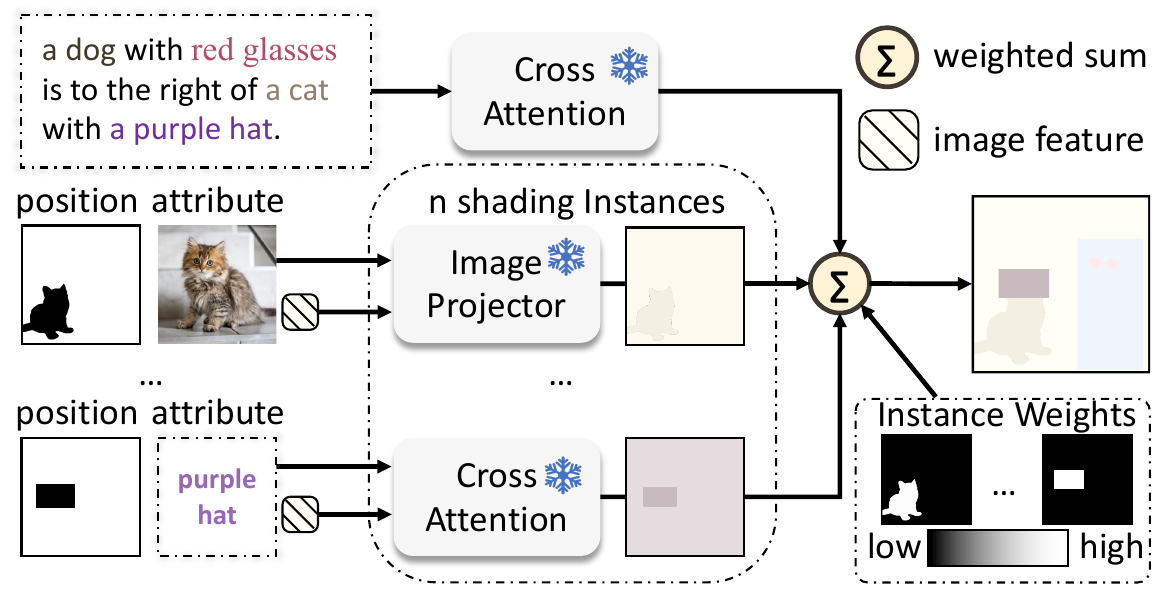}

\vspace{-1.5mm}
    \caption{\textbf{Refined Shader (\S\ref{sec:refined_shader})} uses the pre-trained Cross-Attention layer and image projector to independently shade each instance's details, finally combining them. }
    \label{fig:refinedshader}
    
\vspace{-3.5mm}

\end{figure}

\subsubsection{Refined Shader}\label{sec:refined_shader}

\begin{figure}[tb]
    \centering
    \includegraphics[width=1.0\linewidth]{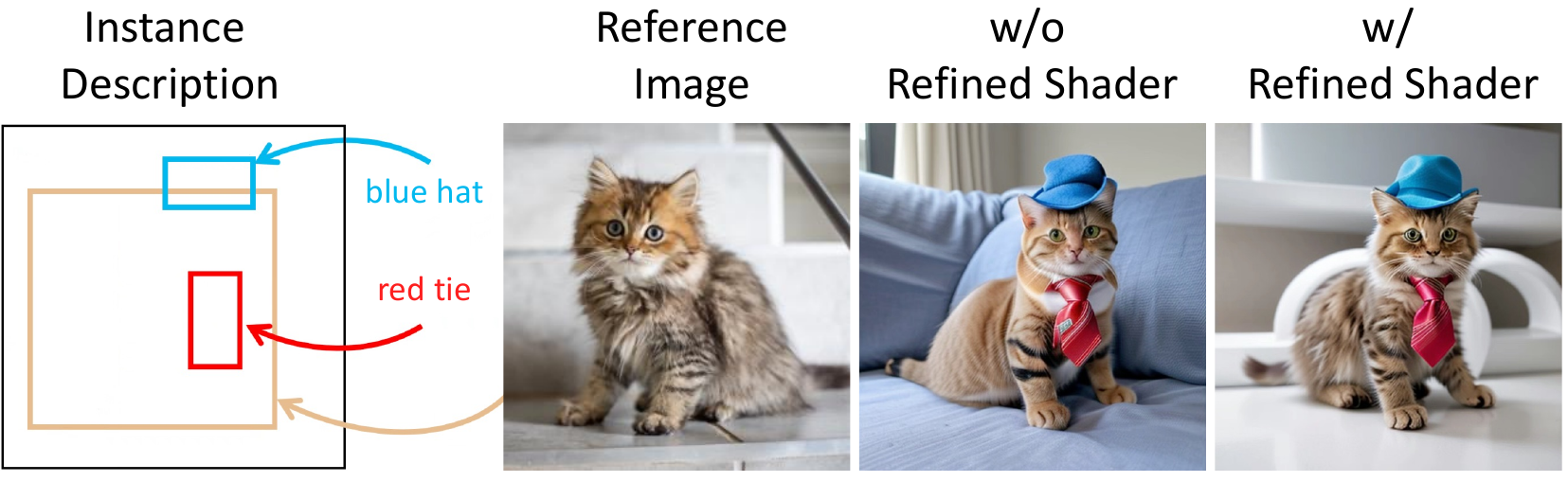}

\vspace{-1.5mm}
    \caption{
    \textbf{Effect of the Refined Shader (\S\ref{sec:refined_shader}).}
    }
    \label{fig:refinedshader_effect}
    
\vspace{-3.5mm}

\end{figure}

\textbf{Motivation.} 
Instead of uniformly applying the Instance Shader (\S\ref{sec:instance_shader}) across all blocks and sample steps, MIGC strategically places it in the mid-blocks and deep up-blocks of the U-net architecture. This placement optimally controls layout and coarse attributes such as category, color, and shape, as shown in Tab.~\ref{tab:instanceshader_pos_ablation}. However, as Fig.~\ref{fig:migc_overview}(a) indicates, some blocks retain standard Cross-Attention layers, leading to detail shading inaccuracies, such as rendering a banana with apple-like details. These discrepancies, critical when precise adherence to a reference image is required, result in instances diverging from their intended appearance (Fig.~\ref{fig:refinedshader_effect}). To resolve this, we propose a training-free Refined Shader to replace all Cross-Attention layers in SD, enhancing the accuracy of multi-instance detail shading.

\noindent\textbf{Solution.} 
As illustrated in Fig.~\ref{fig:refinedshader}, the Refined Shader also operates under a divide-and-conquer framework, shading each instance independently before integration. Given that the Instance Shaders have accurately positioned each instance at its designated location with the correct attributes at a coarse granularity, the Refined Shaders employ pre-trained Cross-Attention layers~\cite{stablediffusion,crossmodal} or Image Projectors~\cite{elite,ssr,ip-adapter} for fine-detail shading, exploiting on the pre-trained model's capability for generating detailed visuals.
Given the image description, the Refined Shader initially employs the pretrained Cross-Attention layer as per Eq.~\ref{eq:cross_text} to derive a global shading result $\textbf{R}^{c}_{ref}$. Then, for each instance $\textbf{i}$ with attribute description ${attr}_{i}$, the Refined Shader gets the shading result per Eq.~\ref{eq:cross_text} and Eq.~\ref{eq:cross_image}:

\setlength{\abovedisplayshortskip}{2pt}
\begin{equation}\small
\label{eq:refined_shading}
\textbf{R}_{ref}^{i} = 
\begin{cases} 
\mathrm{\mathbf{CA}}(\mathbf{X}, {attr}_{i}) & \text{if } {attr}_{i} \in \text{text}, \\
\phantom{I}\mathrm{\mathbf{IP}}(\mathbf{X}, {attr}_{i}) & \text{if } {attr}_{i} \in \text{image}.
\end{cases}
\end{equation}
These results $\{\textbf{R}^{c}_{ref},\textbf{R}^{1}_{ref}, \cdots, \textbf{R}^{n}_{ref}\}$ are then integrated using a weighted sum function. Specifically, a 2D weight map is constructed for each instance. Within the weight map $\textbf{m}_{i}$ of the instance $\boldsymbol{i}$, pixels within the instance's defined region ${pos}_{i}$ are assigned a value of $\boldsymbol{\alpha}$, while all other pixels are set to a value of $\textbf{0}$. In contrast, the weight map $\textbf{m}_{c}$ for the global shading result uniformly assumes a value of $\boldsymbol{\beta}$. A softmax function is applied across these weight maps $\{\textbf{m}_{c},\textbf{m}_{1}, \cdots, \textbf{m}_{n}\}$ in the 2D space to establish their final weights $\{\bar{\textbf{m}}_{c},\bar{\textbf{m}}_{1}, \cdots, \bar{\textbf{m}}_{n}\}$ , which are then used to calculate the aggregate shading result:
\begin{equation}\small
\label{eq:refined_shading_result}
\textbf{R}_{ref}=\bar{\textbf{m}}_{c}\cdot\textbf{R}^{c}_{ref}+\bar{\textbf{m}}_{1}\cdot\textbf{R}^{1}_{ref} + \cdots + \bar{\textbf{m}}_{n}\cdot\textbf{R}^{n}_{ref}.
\end{equation}
Refined shader offers a significant advantage over the vanilla Cross-Attention layer for multi-instance detail shading, as shown in the Fig.~\ref{fig:migc_overview} and Fig.~\ref{fig:refinedshader_effect}.

\subsubsection{Deployment of MIGC++}\label{sec:deployment_migc+}
Fig.~\ref{fig:migc_overview}(b) outlines the deployment strategy of MIGC++. MIGC++ positions the Instance Shader similar to MIGC and equips all blocks with the Refined Shader. For blocks containing both the Instance Shader and the Refined Shader, MIGC++ employs a learned scalar to effectively merge the multi-instance shading results from these two shaders:
\begin{equation}\small
\label{eq:two_shader}
\mathbf{R}_{merge}=\mathbf{R}_{inst}\cdot\mathbf{\mathrm{tanh}}({\boldsymbol{\gamma}}) + \textbf{R}_{ref},
\end{equation}
where the scalar $\boldsymbol{\gamma}$ is inited as \textbf{0}.

\begin{figure}[tb]
    \centering
    \includegraphics[width=1.0\linewidth]{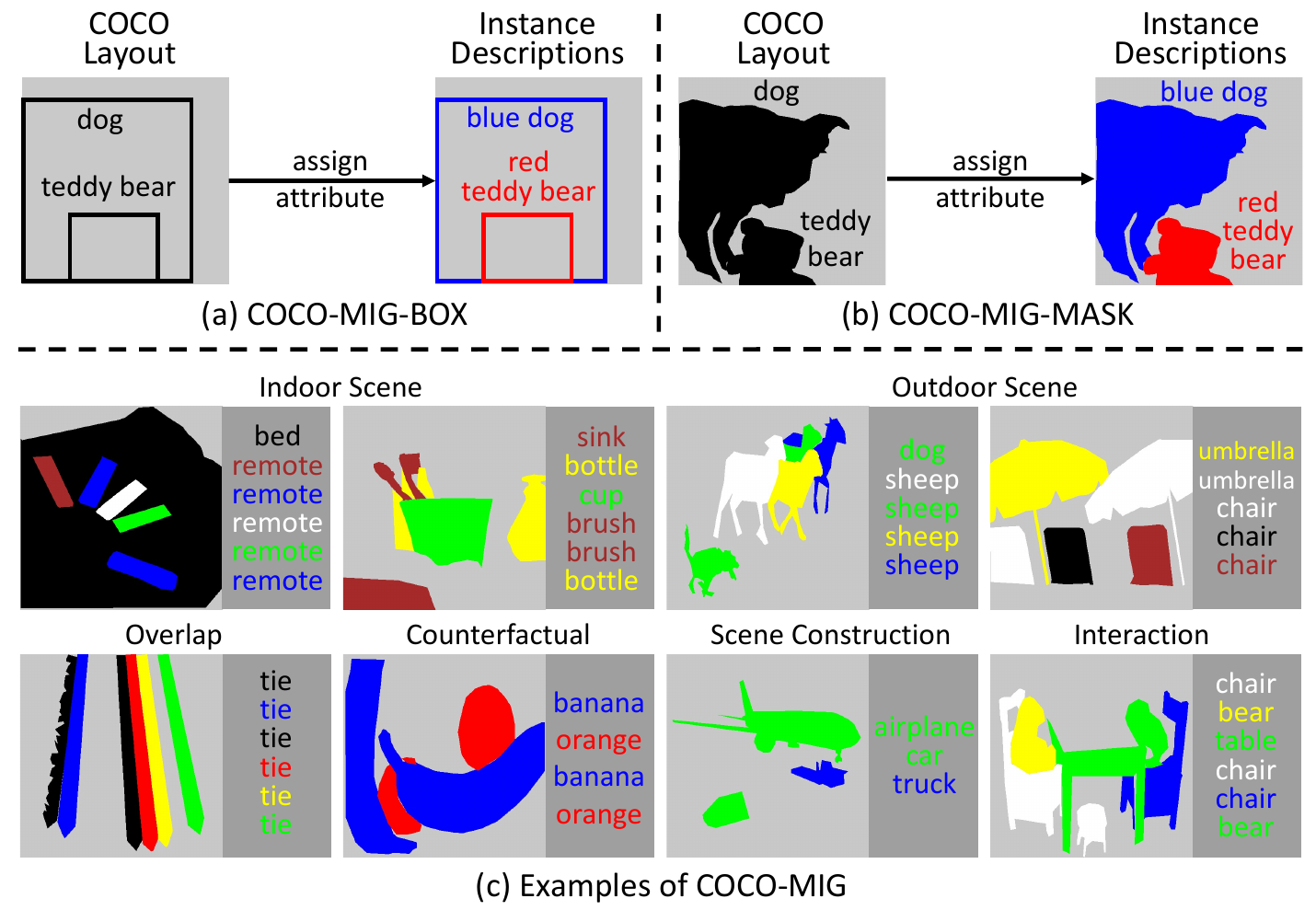}

\vspace{-1.5mm}
    \caption{\textbf{COCO-MIG (\S\ref{sec:cocomig}).} (a) (b) The COCO-MIG benchmark uses sampled layouts from the COCO dataset, assigning a specific color attribute to each instance. (c) This benchmark includes both indoor and outdoor scenes, addressing challenges such as overlap, counterfactual elements, and scene interactions. Color attributes for each instance are visually represented by masks in corresponding colors. }
    \label{fig:cocomig_construct}
    
\vspace{-3.5mm}

\end{figure}

\vspace{-3.0mm}
\subsection{Consistent-MIG}\label{sec:consistent_mig}
The Consistent-MIG algorithm improves the iterative MIG capabilities of MIGC \& MIGC++, facilitating modifying certain instances in MIG while preserving consistency in unmodified regions and maximizing the ID consistency of modified instances, as depicted in Fig.~\ref{fig:mig_overview}(c).

\noindent\textbf{Consistency of Unmodified Areas.} Inspired by Blended Diffusion~\cite{blenddiff}, Consistent-MIG maintains consistency in unmodified areas during iterative MIG by replacing the results of unmodified areas with those from the previous iteration. Specifically, during iterative generation, we obtain a mask $\mathbf{m}_{modify}$ by comparing the differences between the instance descriptions of two iterations, where modified regions are marked as 1 and unmodified regions as 0. 
Assuming the result sampled in the previous iteration is $\mathbf{z}_{t,prev}$, we use the mask $\mathbf{m}_{modify}$ to update the current iteration's sampled result $\mathbf{z}_{t,cur}$ for consistency in the unmodify areas:
\begin{equation}\small
\label{eq:bg_consistent}
\mathbf{z}'_{t,cur} = \mathbf{m}_{modify}\cdot\mathbf{z}_{t,cur} + (1-\mathbf{m}_{modify})\cdot\mathbf{z}_{t,prev}.
\end{equation}

\noindent\textbf{Consistency of Identity.} 
ID consistency is essential when modifying instances' attributes like color. Drawing on techniques~\cite{tuneavideo} from Text-to-Video generation to ensure temporal consistency, our method, Consistent-MIG, employs a specific strategy to ensure ID consistency. In the Self-Attention phase, both the key and value from the previous iteration are concatenated with the current key and value to enhance the utilization of ID information from prior iterations, thus preserving identity continuity.

\begin{table*}[t!]
	\centering
	\caption{\textbf{Quantitative results (\S\ref{sec:compare}) on proposed COCO-MIG-BOX}. $L_{i}$ means that the count of instances needed to generate in the image is \textbf{i}.}\label{tab:coco_mig_box}
	\vspace{-3.5mm}
\centering
\setlength{\tabcolsep}{4.5pt}
\renewcommand\arraystretch{1.1}
\begin{tabular}{rcccccccccccccccccc}
\bottomrule[1pt]\rowcolor[HTML]{FAFAFA}
                                         & \multicolumn{6}{c}{Instance Success Ratio$\uparrow$}    & \multicolumn{6}{c}{Image Success Ratio$\uparrow$} & \multicolumn{6}{c}{Mean Intersection over Union$\uparrow$} \\

\cmidrule(lr){1-1} \cmidrule(lr){2-7} \cmidrule(lr){8-13}  \cmidrule(lr){14-19} 
Method                                  & $\mathcal{L}2$         & $\mathcal{L}3$ & $\mathcal{L}4$ & $\mathcal{L}5$ & $\mathcal{L}6$ & $\mathcal{AVG}$ & $\mathcal{L}2$         & $\mathcal{L}3$ & $\mathcal{L}4$ & $\mathcal{L}5$ & $\mathcal{L}6$ & $\mathcal{AVG}$ & $\mathcal{L}2$         & $\mathcal{L}3$ & $\mathcal{L}4$ & $\mathcal{L}5$ & $\mathcal{L}6$ & $\mathcal{AVG}$         \\ \toprule[0.8pt]
\rowcolor[HTML]{F9FFF9} SD\hspace{0.55em}\pub{CVPR22}\hspace{0.1em}\cite{stablediffusion}         & 6.9 & 3.9                       & 2.5          & 2.7          & 2.4         & 3.2          & 0.3 & 0.0                       & 0.0         & 0.0          & 0.0         & 0.0          & 4.6 & 2.8                       & 1.7          & 1.9          & 1.6          & 2.2                      \\ 
\rowcolor[HTML]{F9FFF9} TFLCG\hspace{0.0em}~\pub{WACV24}\hspace{0.1em}\cite{tflcg}               & 17.2 & 13.5                       & 7.9          & 6.1          & 4.5          & 8.3          & 3.8 & 0.2                       & 0.0          & 0.0          & 0.0          & 0.8 & 10.9 & 8.7                       & 5.1          & 3.9          & 2.8          & 5.3            \\
\rowcolor[HTML]{F9FFF9} BoxDiff\hspace{0.48em}~\pub{ICCV23}\hspace{0.1em}\cite{boxdiff}                                      & 26.5 & 21.4                       & 14.5          & 11.0          & 10.2          & 14.6          & 8.2 & 1.5                       & 0.3          & 0.0          & 0.0          & 2.0 & 18.2 &14.6 & 9.8 & 7.3 & 6.7 & 9.9          \\
\rowcolor[HTML]{F9FFF9} MultiDiff\hspace{0.40em}~\pub{ICML23}\hspace{0.1em}\cite{multidiffusion}             &    28.0        & 24.4                       & 22.1          & 18.2          & 19.4          & 21.3          & 8.3        & 1.9                       & 0.3          & 0.4          & 0.6          & 2.4  & 20.3        & 17.5                       & 15.7          & 13.1          & 13.7          & 15.2        \\
\rowcolor[HTML]{F9FFF9} GLIGEN\hspace{0.30em}~\pub{CVPR23}\hspace{0.1em}\cite{gligen}            & 41.8 & 34.4                       & 31.9          & 27.9          & 29.9          & 31.8          & 16.4 & 2.9                       & 0.9          & 0.3          & 0.0          & 4.2   & 35.0 & 28.2                       & 25.9          & 22.4          & 23.8          & 25.7         \\

\rowcolor[HTML]{F9FFF9} InstanceDiff\hspace{0.3em}~\pub{CVPR24}\hspace{0.1em}\cite{instdiff} & 61.0 & 52.8                       & 52.4          & 45.2          & 48.7          & 50.5          & 36.5 & 15.3                       & 9.8          & 2.7          & 3.6          & 13.8 & 53.8 & 45.8                       & 44.9          & 37.7          & 40.6          & 43.0  \\
\rowcolor[HTML]{F9FFF9} RECO\hspace{0.30em}~\pub{CVPR23}\hspace{0.1em}\cite{reco}                                  & 65.5 & 56.1                       & 56.3          & 52.4          & 58.3          & 56.9          & 40.8 & 19.5                       & 13.8          & 7.3          & 11.2          & 18.8  & 55.7 & 46.7                       & 47.2          & 43.3          & 48.8          & 47.6        \\
\rowcolor[HTML]{F9FBFF} \toprule[0.8pt]
\rowcolor[HTML]{F9FBFF}
MIGC\hspace{0.30em}~\pub{CVPR24}\hspace{0.1em}\cite{migc}                                  & \textbf{76.2} & \underline{70.1} & \underline{70.4} & \underline{66.4} & \underline{67.7} & \underline{69.2} & \textbf{58.5} & \underline{36.3} & \underline{25.9} & \underline{16.2} & \underline{16.6} & \underline{31.0} & \underline{64.4} & \underline{58.5} & \underline{57.6} & \underline{53.6} & \underline{54.2} & \underline{56.5}         \\
\rowcolor[HTML]{F9FBFF}
\multicolumn{1}{c}{MIGC++}                             & \underline{74.6} & \textbf{72.1} & \textbf{72.4} & \textbf{69.0} & \textbf{71.3} & \textbf{71.4} & \underline{55.8} & \textbf{39.7} & \textbf{28.6} & \textbf{19.7} & \textbf{21.6} & \textbf{33.4} & \textbf{65.8} &  \textbf{62.2} & \textbf{61.7} & \textbf{57.5} & \textbf{59.2}  & \textbf{60.4}         \\

\toprule[1pt]
\end{tabular}
\vspace{-4.0mm}
\end{table*}

\begin{table*}[t]
	\centering
	\caption{\textbf{Quantitative results (\S\ref{sec:compare}) on proposed COCO-MIG-MASK}. \textblue{$^\dagger$: Using only box annotation data during training.} 
	}\label{tab:coco_mig_mask}
	\vspace{-3.5mm}
\centering
\setlength{\tabcolsep}{4.5pt}
\renewcommand\arraystretch{1.1}
\begin{tabular}{rcccccccccccccccccc}
\bottomrule[1pt]\rowcolor[HTML]{FAFAFA}
                                         & \multicolumn{6}{c}{Instance Success Ratio$\uparrow$}    & \multicolumn{6}{c}{Image Success Ratio$\uparrow$} & \multicolumn{6}{c}{Mean Intersection over Union$\uparrow$} \\

\cmidrule(lr){1-1} \cmidrule(lr){2-7} \cmidrule(lr){8-13}  \cmidrule(lr){14-19} 
\multicolumn{1}{c}{Method}                                  & $\mathcal{L}2$         & $\mathcal{L}3$ & $\mathcal{L}4$ & $\mathcal{L}5$ & $\mathcal{L}6$ & $\mathcal{AVG}$ & $\mathcal{L}2$         & $\mathcal{L}3$ & $\mathcal{L}4$ & $\mathcal{L}5$ & $\mathcal{L}6$ & $\mathcal{AVG}$ & $\mathcal{L}2$         & $\mathcal{L}3$ & $\mathcal{L}4$ & $\mathcal{L}5$ & $\mathcal{L}6$ & $\mathcal{AVG}$         \\ \toprule[0.8pt]
\rowcolor[HTML]{F9FFF9} SD\hspace{0.55em}\pub{CVPR22}\hspace{0.1em}\cite{stablediffusion}         & 2.7 & 1.0 & 0.6 & 0.6 & 0.7 & 0.9 & 0.0 & 0.0 & 0.0 & 0.0 & 0.0 & 0.0 & 1.6 & 0.6 & 0.4 & 0.4 & 0.4 & 0.6                     \\ 
\rowcolor[HTML]{F9FFF9} TFLCG\hspace{0.0em}~\pub{WACV24}\hspace{0.1em}\cite{tflcg}               & 11.1 & 6.0 & 3.9 & 2.4 & 1.7 & 3.9 & 2.3 & 0.2 & 0.0 & 0.0 & 0.0 & 0.0 & 6.8 & 3.7 & 2.4 & 1.5 & 1.0 & 2.4            \\
\rowcolor[HTML]{F9FFF9} MultiDiff\hspace{0.40em}~\pub{ICML23}\hspace{0.1em}\cite{multidiffusion}             & 20.9 & 18.7 & 15.5 & 12.9 & 13.6 & 15.4 & 8.3 & 1.9 & 0.3 & 0.4 & 0.6 & 2.4 & 14.3 & 12.9 & 10.6 & 8.8 & 9.2 & 10.5        \\

\rowcolor[HTML]{F9FFF9} InstanceDiff\hspace{0.3em}~\pub{CVPR24}\hspace{0.1em}\cite{instdiff} & \underline{65.3} & {53.2} &  {53.9} & {45.7} & {48.8} & {51.5}  & \underline{42.8} & {17.5} & {11.5} & {3.9} & {4.5} & {16.3} & \underline{52.7} & \underline{42.6} & \underline{43.5} & \underline{36.2} & \underline{39.9} & \underline{41.5}  \\

\toprule[0.8pt]
\rowcolor[HTML]{F9FBFF} 
\textblue{MIGC$^\dagger$}  \hspace{0.30em}~\pub{CVPR24}\hspace{0.1em}\cite{migc}                          & \textblue{62.3} & \textblue{\underline{55.8}} & \textblue{\underline{56.3}} & \textblue{\underline{47.7}} & \textblue{\underline{51.6}} & \textblue{\underline{53.4}} & 

\textblue{40.6} & \textblue{\underline{19.6}} & \textblue{\underline{12.2}} & \textblue{\underline{5.7}} & \textblue{\underline{5.2}} & \textblue{\underline{16.9}} & 

\textblue{44.7} & \textblue{38.9} & \textblue{39.7} & \textblue{33.0} & \textblue{36.6} & \textblue{37.6} \\

\rowcolor[HTML]{F9FBFF} \multicolumn{1}{c}{MIGC++}                             & \textbf{73.8} & \textbf{68.6} & \textbf{68.8} & \textbf{62.9} & \textbf{67.6} & \textbf{67.5} & \textbf{53.2} & \textbf{36.0} & \textbf{23.8} & \textbf{11.6} & \textbf{17.4} & \textbf{28.8} & \textbf{61.9} & \textbf{56.7} & \textbf{56.2} & \textbf{50.2} & \textbf{54.2} & \textbf{54.8}          \\

\toprule[1pt]
\end{tabular}
\vspace{-4.0mm}
\end{table*}

\begin{table}[t!]
	\centering
	\caption{\textbf{Quantitative results (\S\ref{sec:compare}) on COCO-Position}. 
	}\label{tab:coco_pos}
	\vspace{-3.5mm}
\centering
\setlength{\tabcolsep}{2.5pt}
\renewcommand\arraystretch{1.1}
\begin{tabular}{rcccccccccc}
\bottomrule[1pt]\rowcolor[HTML]{FAFAFA}
                                         & \multicolumn{3}{c}{Position Accuracy$\uparrow$}    & \multicolumn{2}{c}{CLIP Score$\uparrow$} \\

\cmidrule(lr){1-1} \cmidrule(lr){2-4} \cmidrule(lr){5-6}  
\multicolumn{1}{c}{Method}                                  & $SR$         & $MIoU$ & $AP$ &  $global$ & $local$         & $FID$$\downarrow$         \\ 
\toprule[0.8pt]
\rowcolor[HTML]{F5F5F5}
\multicolumn{1}{c}{Real Image}         & 83.75 & 85.49 & 65.97  & 24.22 & 19.74 & -                 \\ 
\toprule[0.8pt]
\rowcolor[HTML]{F9FFF9} SD\hspace{0.55em}\pub{CVPR22}\hspace{0.1em}\cite{stablediffusion}         & 5.95 & 21.60 & 0.80  & \textbf{25.69} & 17.34 & \textbf{23.56}                \\ 
\rowcolor[HTML]{F9FFF9} TFLCG\hspace{0.0em}~\pub{WACV24}\hspace{0.1em}\cite{tflcg}               & 13.54 & 28.01 & 1.75 & \underline{25.07} & 17.97 & 24.65            \\
\rowcolor[HTML]{F9FFF9} BOXDiff\hspace{0.4em}~\pub{ICCV23}\hspace{0.1em}\cite{tflcg}               & 17.84 & 33.38 & 3.29 & 23.79 & 18.70 & 25.15            \\
\rowcolor[HTML]{F9FFF9} MultiDiff\hspace{0.36em}~\pub{ICML23}\hspace{0.1em}\cite{multidiffusion}             & 23.86 & 38.82 & 6.72 & 22.10 & 19.13 & 33.20         \\

\rowcolor[HTML]{F9FFF9} LayoutDiff\hspace{0.4em}~\pub{ICCV23}\hspace{0.1em}\cite{multidiffusion}             & 50.53 & 57.49 & 23.45& 18.28 & 19.08 & 25.94         \\

\rowcolor[HTML]{F9FFF9} GLIGEN\hspace{0.30em}~\pub{CVPR23}\hspace{0.1em}\cite{instdiff} & 70.52 & 71.61 & 40.68 & 24.61 & 19.69 & 26.80  \\

\toprule[0.8pt]

\rowcolor[HTML]{F9FBFF} MIGC\hspace{0.30em}~\pub{CVPR24}\hspace{0.1em}\cite{migc}                             & \underline{80.29} & \underline{77.38} & \underline{54.69} & 24.66 & \underline{20.25} & 24.52           \\

\rowcolor[HTML]{F9FBFF} \multicolumn{1}{c}{MIGC++} & \textbf{82.54}  & \textbf{81.02} & \textbf{63.56} & 24.71 & \textbf{20.85}  & \underline{24.42}        \\

\toprule[1pt]
\end{tabular}
\vspace{-4.0mm}
\end{table}

\vspace{-3.0mm}
\subsection{Loss Function}\label{sec:loss}
Given the image description $\boldsymbol{c}$ and instance descriptions $\mathbb{I}$, we utilize the original denoising loss of the SD to train the Instance Shader $\boldsymbol{\theta}^{\prime}$, keeping the parameters $\boldsymbol{{\theta}}$ of the SD model frozen during this process:
\begin{equation}\small
\label{eq:ori_loss}
\min _{\boldsymbol{\theta^{\prime}}} \mathcal{L}_{\mathrm{ldm}}=\mathbb{E}_{\boldsymbol{z}, \boldsymbol{\epsilon} \sim \mathcal{N}(\mathbf{0}, \mathbf{I}), t}\left[\left\|\boldsymbol{\epsilon}-f_{\boldsymbol{\theta, \theta^{\prime}}}\left(\boldsymbol{z}_t, t, \boldsymbol{c},\mathbb{I}\right)\right\|_2^2\right].
\end{equation}

Furthermore, to ensure that the \textbf{n} generated instances remain within their designated position and to prevent the inadvertent creation of extraneous instances in the background, we have developed an inhibition loss. This loss function is specifically designed to avoid high attention weights in the background areas:
\begin{equation}\small
\label{eq:inhb_loss}
\min _{\boldsymbol{\theta^{\prime}}} \mathcal{L}_{\mathrm{ihbt}}=\scriptstyle{\sum_{i=1}^n}\left|\mathbf{A}_{c, \theta, \theta^{\prime}}^i-\mathrm{\mathbf{DNR}}\left(\mathbf{A}_{c,\theta,\theta^{\prime}}^i\right)\right| \odot \mathbf{M}^{b g},
\end{equation}
where $\mathbf{A}_{c, \theta, \theta^{\prime}}^i$ represents the attention map for the ith instance, derived from the Cross-Attention Layers that are frozen in the proposed Refined Shader model within the 16x16 decoder blocks, and DNR(·) represents the denoising of the background region, achieved using an averaging operation. The design of the final training loss is as follows:
\begin{equation}\small
\label{eq:loss}
\mathcal{L}=\mathcal{L}_{ldm}+\boldsymbol{\gamma}\cdot\mathcal{L}_{ihbt},
\end{equation}
where we set the weight $\boldsymbol{\gamma}$ as 0.1.

\vspace{-3.0mm}
\subsection{Implementation Details}\label{sec:implementation}

\noindent\textbf{Data Preparation.} We train MIGC and MIGC++ on COCO 2014~\cite{coco}, with each image having an annotated image description. We employ the Stanza~\cite{stanza} to parse the instance textual descriptions within the image descriptions. Based on the instance textual description, we use Grounding-DINO~\cite{gdino} to detect each instance's bounding box and further employ Grounded-SAM~\cite{sam} to obtain the mask of each instance. Using the bounding box of each instance, we crop out each instance as an instance image description.

\noindent\textbf{Training Text Modality.} To put the data in the same batch, we set the instance count to \textbf{6} during training. If an image contains more than \textbf{6} instances, \textbf{6} of them will be randomly selected. If the data contains fewer than \textbf{6} instances, we complete it with null text, specifying the position with a bounding box of [0.0, 0.0, 0.0, 0.0] or an all-zero mask. We train our MIGC and MIGC++ based on the pre-trained SD1.4~\cite{stablediffusion}. We use the AdamW \cite{adam} optimizer with a constant learning rate of $1e^{-4}$ and a decay weight of $1e^{-2}$, training the model for 300 epochs with a batch size of 320.

\noindent\textbf{Training Image Modality.} We froze the pre-trained weights in the text modality and introduced a new Enhance-Attention layer for the Image modality, enabling MIGC++ to perform shading based on the reference image. Consistent with the training of the text modality, in order to put the data in one batch, we fix the number of text-described instances to \textbf{6} and the number of image-described instances to \textbf{4} during training. If there are fewer than \textbf{4} instances in the data, we pad the image-described instances with a blank white image. We use the AdamW optimizer with a constant learning rate of $1e^{-4}$ and a decay weight of $1e^{-2}$, training the Enhance-Attention layer of the image modality for 200 epochs with a batch size of 312.

\noindent\textbf{Inference.} We use the EulerDiscreteScheduler~\cite{elud} with 50 sample steps. We select the CFG~\cite{Ho2022ClassifierFreeDG} scale as 7.5. For instances described by images, we mask them out, i.e., replace the background area with a blank. We use the ELITE~\cite{elite} as the image projector. The deployment details for MIGC and MIGC++ can be derived from (\S\ref{sec:migc_deployment}) and (\S\ref{sec:deployment_migc+}).


\section{MIG Benchmark}\label{sec:MIG_bench}


\begin{figure*}[tb]
    \centering
    \includegraphics[width=1.0\linewidth]{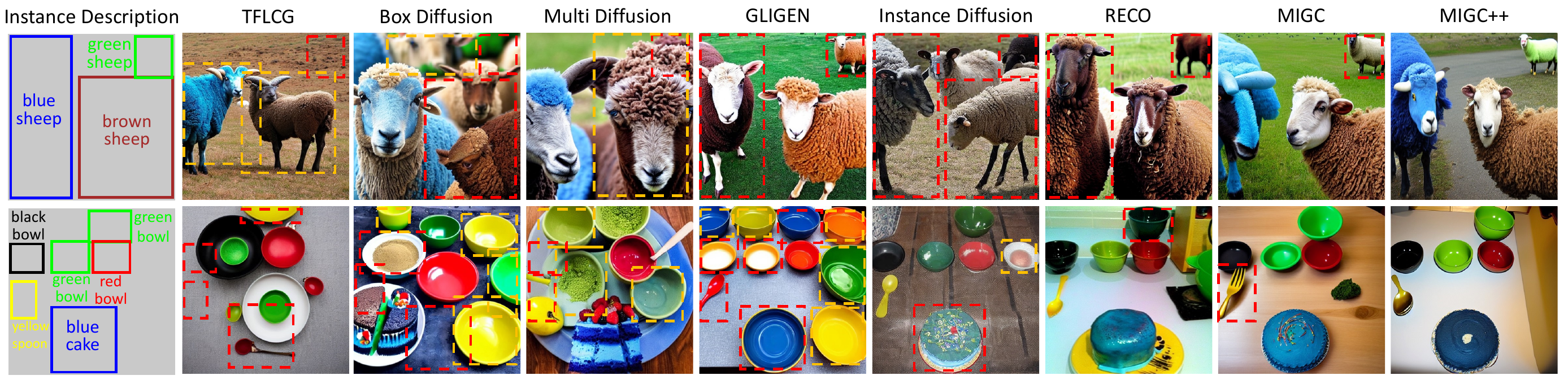}

\vspace{-1.5mm}
    \caption{\textbf{Qualitative results (\S\ref{sec:compare}) on proposed COCO-MIG-BOX}. \textcolor{red}{Red boxes} indicate instances that are missing or have incorrect attributes. \textcolor{yellow}{Yellow boxes} are used to indicate instances that are not generated precisely in their specified locations.}
    \label{fig:cocomig_box_vis}
\vspace{-5mm}
\end{figure*}
\begin{figure}[tb]
    \centering
    \includegraphics[width=1.0\linewidth]{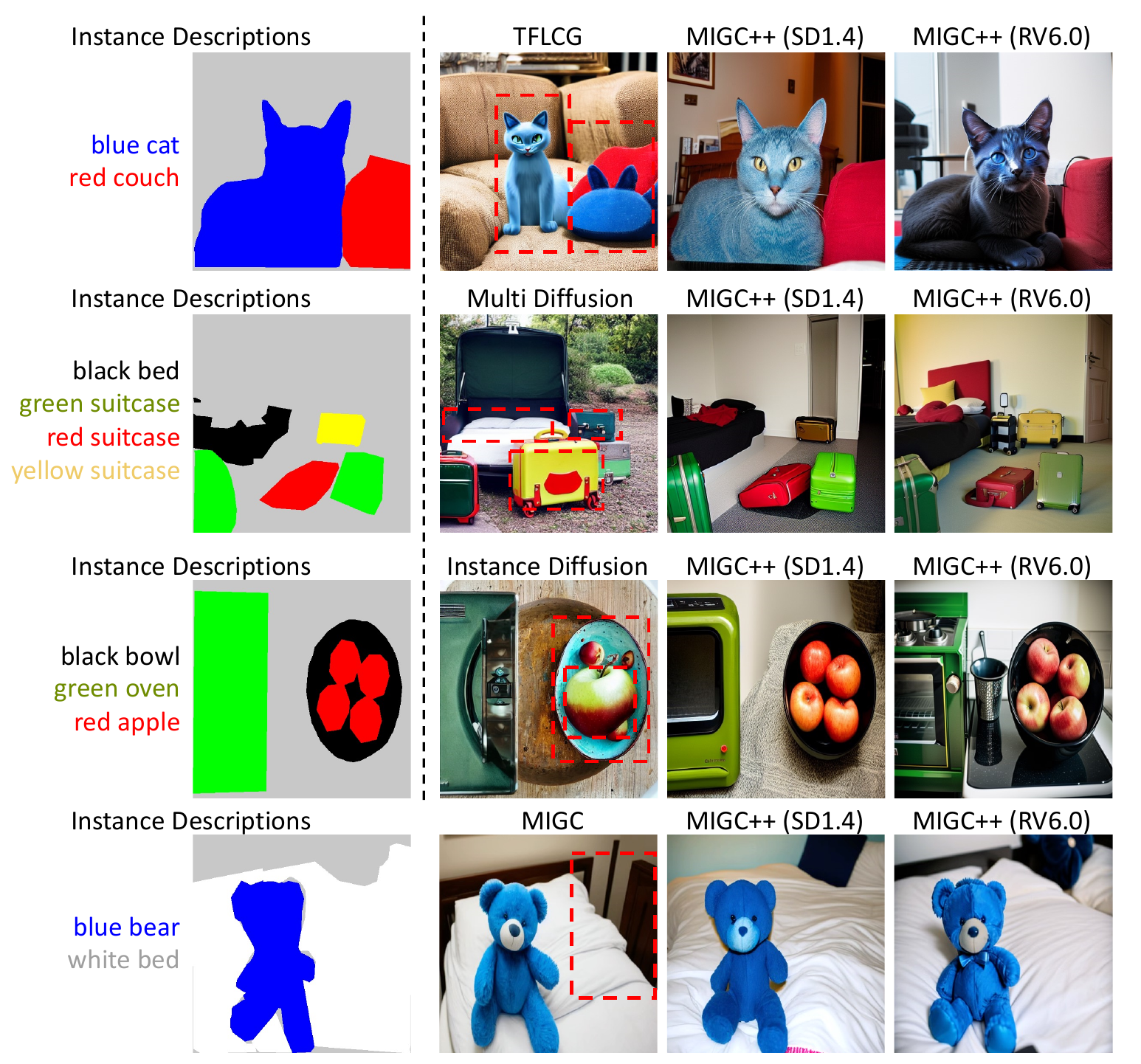}

\vspace{-1.5mm}
    \caption{\textbf{\textblue{Qualitative results (\S\ref{sec:compare}) on proposed COCO-MIG-MASK}}. The \textcolor{red}{red boxes} highlight instances that were not generated according to the mask or have incorrect attributes.}
    \label{fig:cocomig_mask_vis}
    
\vspace{-3.5mm}

\end{figure}

\subsection{COCO-MIG Benchmark}\label{sec:cocomig}
\noindent\textbf{Overview.} 
We introduce the COCO-MIG Benchmark to study the Multi-Instance Generation (MIG) task. To obtain the instance descriptions needed in MIG, i.e., the position and attribute of instances, we sample the layout from the COCO dataset~\cite{coco} to determine the position of each instance and assign a color to each instance to determine its attribute. This benchmark stipulates that each generated instance must conform to predefined positions and attributes.
We refine the benchmark into two variants based on instance positioning methods: COCO-MIG-BOX and COCO-MIG-MASK, as illustrated in Fig.~\ref{fig:cocomig_construct}.

\noindent\textbf{Construction Process.} We extract layouts from the COCO dataset~\cite{coco}, excluding instances with side lengths less than one-eighth of the image size and layouts containing fewer than two instances. To evaluate the model's capacity to modulate the number of instances, we categorize these layouts into five levels, ranging from $L_2$ to $L_6$. Each level, $L_i$, corresponds to a target generation of \textbf{i} instances per image. The distribution of layouts across these levels includes \textbf{155} for $L_2$, \textbf{153} for $L_3$, \textbf{148} for $L_4$, \textbf{140} for $L_5$, and \textbf{154} for $L_6$, totaling \textbf{750} layouts. When sampling layouts for a specific level $L_i$, if the number of instances exceeds $i$, we select the $i$ instances with the largest areas. Conversely, if the number of instances falls short of $i$, we undertake a resampling process. For each layout, we assign a specific color to each instance from a palette of seven colors: red, yellow, green, blue, white, black, and brown. Additionally, we construct the image description as ``a $<$attr1$>$ $<$obj1$>$, a $<$attr2$>$ $<$obj2$>$, ..., and a ...''. With each prompt generating 8 images, each method produces \textbf{6000} images for comparison.

\noindent\textbf{Evaluation Process.}
We use the following pipeline to determine if an instance has been correctly generated: (1) We start by detecting the instance's bounding box in the generated image using Grounding-DINO~\cite{gdino}, and then compute the Intersection over Union (IoU) with the target bounding box. An instance is considered \textbf{Position Correctly Generated} if the IoU is \textbf{0.5} or higher. If multiple bounding boxes are detected, we select the one closest to the target bounding box for IoU computation. If the instance's position is indicated by a mask, we confirm its correct positional generation by checking if the mask's IoU exceeds \textbf{0.5}. (2) For an instance verified as \textbf{Position Correctly Generated}, we next evaluate its color accuracy using Grounded-SAM~\cite{sam} to segment the instance's area in the image, defined as M. We then calculate the proportion of M matching the specified color in the HSV space, denoted as O. If the O/M ratio is \textbf{0.2} or more, the instance is deemed \textbf{Fully Correctly Generated}.

\noindent{\textbf{Metric.}}
The evaluation metrics for COCO-MIG include: (1) \textit{Instance Success Ratio}, which measures the proportion of instances that are Fully Correctly Generated (see evaluation process in \S\ref{sec:cocomig}). (2) \textit{Image Success Ratio}, measuring the proportion of generated images in which all instances are Fully Correctly Generated. (3) \textit{Mean Intersection over Union (MIoU)}, quantifying the alignment between the actual and target positions of each positioned-and-attributed instance. It is important to note that the MIoU for any instance not Fully Correctly Generated is recorded as zero.

\begin{figure}[tb]
    \centering
    \includegraphics[width=1.0\linewidth]{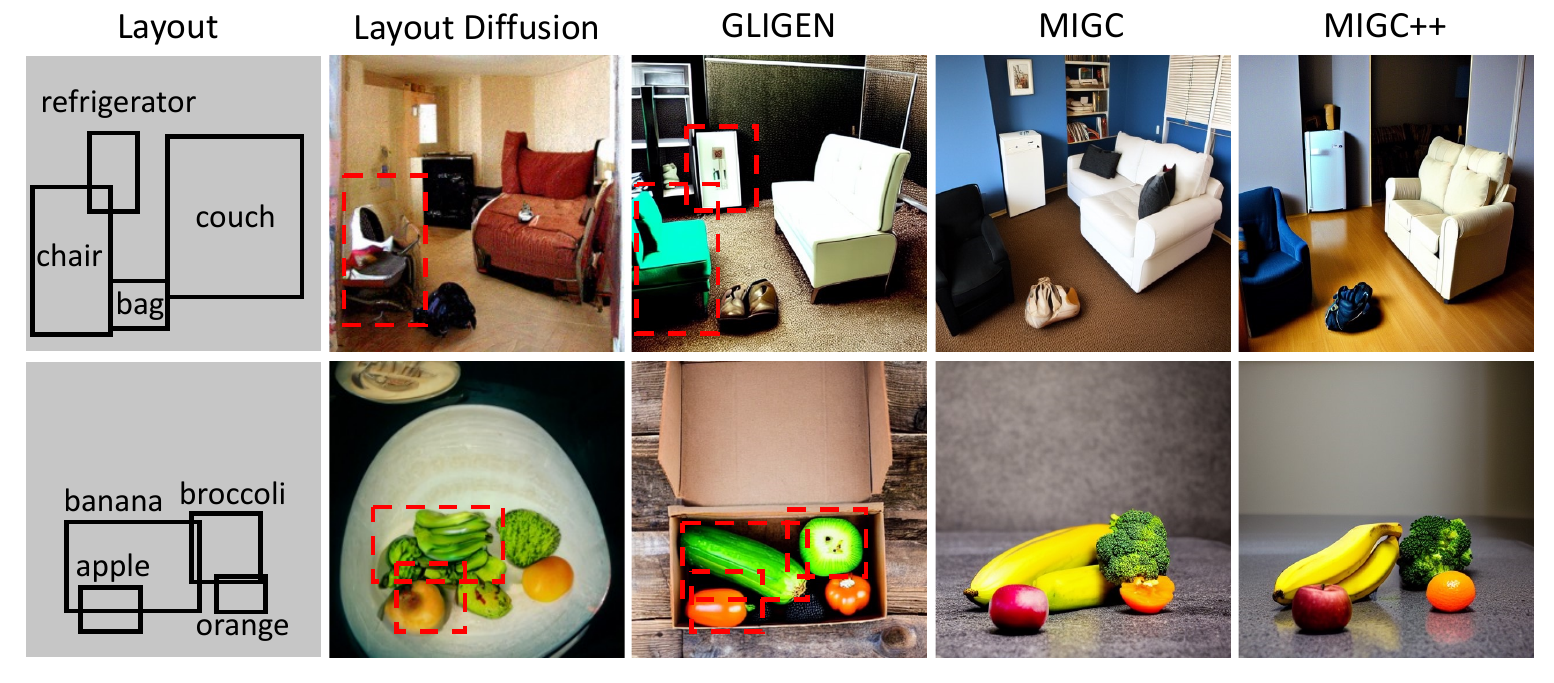}

\vspace{-3.5mm}
    \caption{\textbf{Qualitative results (\S\ref{sec:compare}) on the COCO-Position}. \textcolor{red}{Red boxes} indicate instances that have incorrect category.}
    \label{fig:coco_pos_vis}
    
\vspace{-3.5mm}

\end{figure}

\vspace{-3.0mm}
\subsection{Multimodal-MIG Benchmark}\label{sec:multimodalmig_bench}

\begin{figure}[tb]
    \centering
    \includegraphics[width=1.0\linewidth]{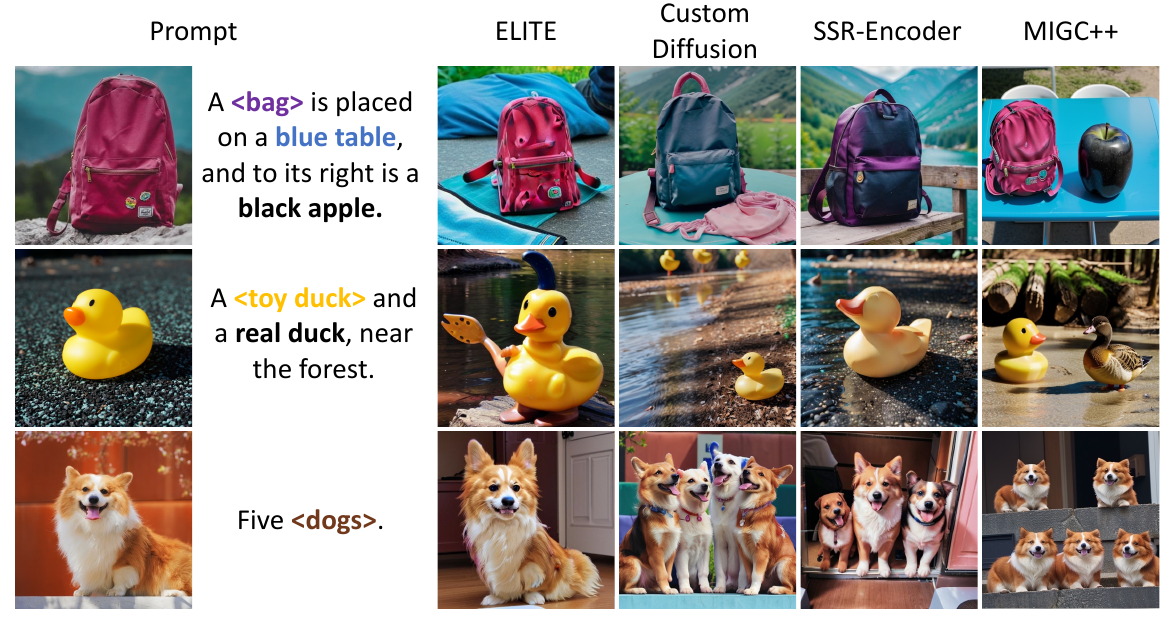}

\vspace{-3.5mm}
    \caption{\textbf{Qualitative Results of the Multimodal-MIG Benchmark.} When utilizing both text and images to describe instances, MIGC++ demonstrates superior capabilities in controlling position, attributes, and quantity.}
    \label{fig:multimodal_mig_vis}
    
\vspace{-3.5mm}

\end{figure}

To evaluate multimodal control over text and image in MIG, we introduce the Multimodal-MIG benchmark. Inspired by DrawBench~\cite{imagen}, this benchmark requires models to manage position, attributes, and quantity of instances, and to align some instances with attributes from a reference image. For example, as shown in the first row of Fig.~\ref{fig:multimodal_mig_vis}, a prompt might use a reference image to specify a ``bag's'' customized attributes and text descriptions such as ``blue table'' and ``black apple'' for other instances. We designed \textbf{20} prompts for this benchmark, with each generating \textbf{10} images, totaling \textbf{200} images per method for evaluation. We employ GPT-4~~\cite{layoutgpt,gpt4} to create instance descriptions necessary for MIGC++, forming an automated two-stage pipeline.


\begin{table}[t!]

\centering
\caption{\textbf{Quantitative results (\S\ref{sec:compare}) on the Drawbench.}}
	\vspace{-3.5mm}
\setlength\tabcolsep{3.5pt}
  \renewcommand\arraystretch{1.1}
  
  \begin{tabular}{r c c | c c | c c  }
    \toprule
    \multicolumn{1}{c}{\multirow{2}{*}{\textbf{Method}}} & \multicolumn{2}{c}{\textbf{Position(\%) $\uparrow$}}  & \multicolumn{2}{c}{\textbf{Attribute(\%) $\uparrow$}}  & \multicolumn{2}{c}{\textbf{Count(\%) $\uparrow$}} \\

    \cmidrule(lr){2-3} \cmidrule(lr){4-5} \cmidrule(lr){6-7} 
    
    & $A$  & $M$ & $A$  & $M$ & $A$  & $M$  \\
    \midrule
    SD\hspace{0.25em}\pub{CVPR22}\hspace{0.1em}~\cite{stablediffusion} & - & 13.3 & - & 57.5 & - & 23.7 \\
    AAE\hspace{0.45em}\pub{SIGG23}\hspace{0.1em}~\cite{aae} & - & 23.1 & - & 51.5 & - & 30.9 \\
    StrucDiff\hspace{0.55em}\pub{ICLR23}\hspace{0.1em}~\cite{structurediff} & - & 13.1 & - & 56.5 & - & 30.3 \\
    \midrule
\rowcolor[HTML]{F9FFF9}    BoxDiff\hspace{0.4em}\pub{ICCV23}\hspace{0.1em}~\cite{boxdiff} & 11.9 & 50.0 & 28.5 & 57.5 & 9.2 & 39.5   \\
\rowcolor[HTML]{F9FFF9}    TFLCG\hspace{0.0em}\pub{WACV24}\hspace{0.1em}~\cite{tflcg} & 9.4 & 53.1 & 35.0 & 60.0 & 15.8 & 31.6    \\
\rowcolor[HTML]{F9FFF9}    MultiDiff\hspace{0.36em}\pub{ICML23}\hspace{0.1em}~\cite{multidiffusion} & 10.6 & 55.6 & 18.5 & 65.5 & 17.8 & 36.2    \\
\rowcolor[HTML]{F9FFF9}    GLIGEN\hspace{0.30em}\pub{CVPR23}\hspace{0.1em}~\cite{gligen} & 61.3 & 78.8 & 51.0 & 48.2 & 44.1 & 55.9  \\
    
    \midrule
\rowcolor[HTML]{F9FBFF}    MIGC\hspace{0.30em}\pub{CVPR24}\hspace{0.1em}~\cite{migc} & \underline{69.4} & \underline{93.1} & \underline{79.0} & \underline{97.5} & \underline{67.8} & \underline{67.5}
    \\
\rowcolor[HTML]{F9FBFF}     \multicolumn{1}{c}{MIGC++} & \textbf{73.8} & \textbf{93.1} & \textbf{82.0} & \textbf{98.5} & \textbf{72.4} & \textbf{71.7}
    \\
    \bottomrule
  \end{tabular}
  \vspace{-1em}
  \label{tab:DrawBench}
\end{table}

\section{Experimental Results}\label{sec:experiments}

We assessed MIGC/MIGC++ using our COCO-MIG and Multimodal-MIG benchmarks and other established ones:

\begin{itemize}[leftmargin=*, noitemsep, topsep=0pt]
\item \textbf{COCO-Position}~\cite{coco} benchmark randomly samples \textbf{800} layouts of realistic scenes from the COCO dataset~\cite{coco} and challenges generative models to create images where each instance adheres to predefined spatial constraints. This test rigorously assesses model precision in instance positioning, resulting in 6,400 images across all layouts. \textbf{Metrics} for COCO-Position include: (1) \textit{Success Ratio (SR)}, measuring correct instance placement; (2) \textit{Mean Intersection over Union (MIoU)}, quantifying positional alignment; (3) \textit{Average Precision (AP)}, detecting extraneous instances; (4) \textit{Global and Local CLIP scores}, evaluating overall and instance-specific alignment of text and image; (5) \textit{FID}, assessing image quality.

\item \textbf{Drawbench}~\cite{imagen} represents a rigorous text-to-image (T2I) benchmark, which is used to measure the capability of image generation models in understanding and converting text descriptions into visual representations. As this benchmark only contains the image description, we employed GPT-4~\cite{gpt4,layoutgpt} to generate instance descriptions for each test prompt. 
With each prompt generating 8 images, each method produces \textbf{512} images for comparison.

\textbf{Metrics} for evaluating DrawBench encompass two main approaches: 
(1) \textit{Automatic evaluation accuracy (A)}, which measures the proportion of generated images where instances are all correctly generated.
(2) \textit{Manual evaluation accuracy (M)}, 
involving ten evaluators who assess whether each image is correctly generated

\end{itemize}
\begin{figure}[t]
    \centering
    \includegraphics[width=1.0\linewidth]{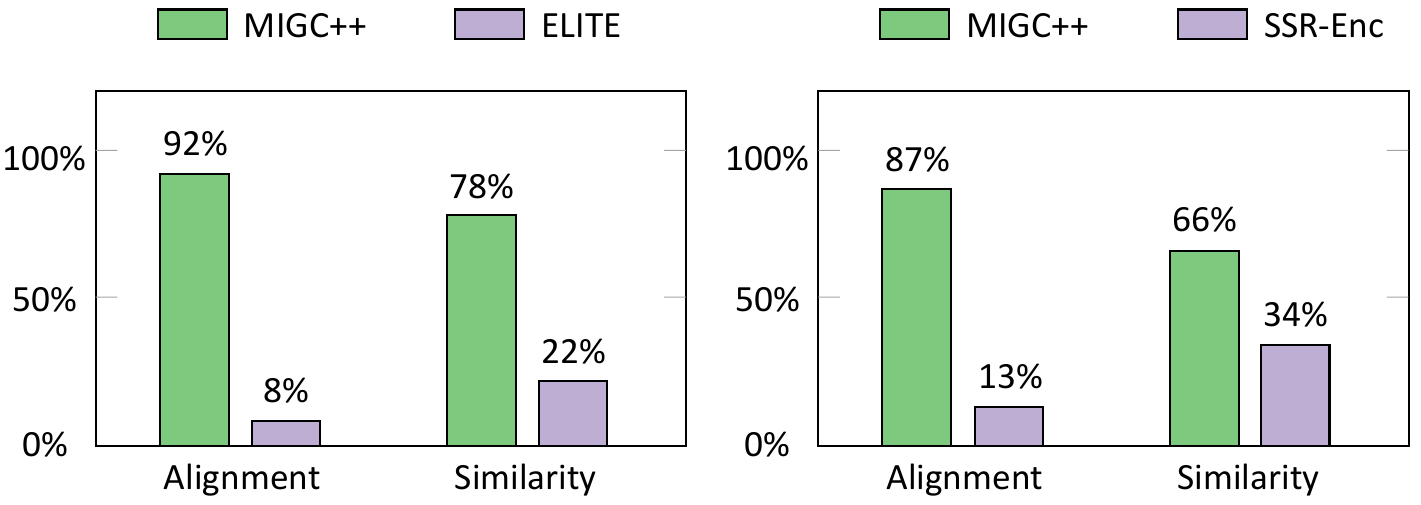}
\vspace{-5.5mm}
    \caption{\textbf{User study on Multimodal-MIG benchmark.}}
    \label{fig:multimodal_mig_user}
    
\vspace{-1.5mm}

\end{figure}
\begin{table}[!t]
\begin{center}

\caption{\textbf{Ablation Study of the Instance Shader Deployment in U-net.} U-net~\cite{Unet,stablediffusion} includes 3 Cross Attention down-blocks, labeled as down-0, down-1, and down-2 from shallow to deep layers. Similarly, there are three up-blocks, labeled as up-1, up-2, and up-3 from deep to shallow layers.}\label{tab:instanceshader_pos_ablation}\vspace{-2mm}
\renewcommand\arraystretch{1.2}
\setlength{\tabcolsep}{3.5pt}
\begin{tabular}{|ccc|c|ccc||ccc|ccc|}
\hline \thickhline
\rowcolor[HTML]{F5F5F5}                      \multicolumn{3}{|c|}{\cellcolor[HTML]{F5F5F5}down}      & \multicolumn{1}{c|}{\cellcolor[HTML]{F5F5F5}mid}                                        & \multicolumn{3}{c||}{\cellcolor[HTML]{F5F5F5}up}
                   & \multicolumn{3}{c|}{\cellcolor[HTML]{F5F5F5}COCO-MIG-BOX}  
                   & \multicolumn{3}{c|}{\cellcolor[HTML]{F5F5F5}COCO-POSITION} 
\\ 
\cline{1-13}  \thickhline

\rowcolor[HTML]{F5F5F5} 
 \multicolumn{1}{|c|}{\cellcolor[HTML]{F5F5F5}0} & \multicolumn{1}{c|}{\cellcolor[HTML]{F5F5F5}1} & 2  & \multicolumn{1}{c|}{\cellcolor[HTML]{F5F5F5}0}      & \multicolumn{1}{c|}{\cellcolor[HTML]{F5F5F5}1} & \multicolumn{1}{c|}{\cellcolor[HTML]{F5F5F5}2} & 3 & \multicolumn{1}{c|}{\cellcolor[HTML]{F5F5F5}ISR} & \multicolumn{1}{c|}{\cellcolor[HTML]{F5F5F5}SR}  & \multicolumn{1}{c|}{\cellcolor[HTML]{F5F5F5}MIoU}  & \multicolumn{1}{c|}{\cellcolor[HTML]{F5F5F5}SR} & \multicolumn{1}{c|}{\cellcolor[HTML]{F5F5F5}AP} & \multicolumn{1}{c|}{\cellcolor[HTML]{F5F5F5}MIoU}                   \\ 
  \hline\hline

\multicolumn{1}{|c|}{}                 & \multicolumn{1}{c|}{}                           &           & \multicolumn{1}{c|}{{}}         &  \multicolumn{1}{c|}{}         & \multicolumn{1}{c|}{}  & & \multicolumn{1}{c|}{5.8}  & \multicolumn{1}{c|}{0.0}  & \multicolumn{1}{c|}{3.6} & \multicolumn{1}{c|}{7.7}  & \multicolumn{1}{c|}{1.4}  & \multicolumn{1}{c|}{23.4} 
\\
 \cline{1-13} 

\multicolumn{1}{|c|}{}                 & \multicolumn{1}{c|}{}                           &           & \multicolumn{1}{c|}{{\checkmark}}         &  \multicolumn{1}{c|}{}         & \multicolumn{1}{c|}{}  & & \multicolumn{1}{c|}{46.9}  & \multicolumn{1}{c|}{12.4}  & \multicolumn{1}{c|}{35.0} & \multicolumn{1}{c|}{65.1}  & \multicolumn{1}{c|}{31.3}  & \multicolumn{1}{c|}{66.0}        \\ 
 \cline{1-13} 

\multicolumn{1}{|c|}{}                 & \multicolumn{1}{c|}{}                           &           & \multicolumn{1}{c|}{}         &  \multicolumn{1}{c|}{{\checkmark}}         & \multicolumn{1}{c|}{}  & & \multicolumn{1}{c|}{\underline{68.4}}  & \multicolumn{1}{c|}{\underline{30.3}}  & \multicolumn{1}{c|}{57.7} & \multicolumn{1}{c|}{\textbf{83.5}}  & \multicolumn{1}{c|}{62.1}  & \multicolumn{1}{c|}{80.7}        \\ 
 \cline{1-13} 

\multicolumn{1}{|c|}{}                 & \multicolumn{1}{c|}{}                           &           & \multicolumn{1}{c|}{\checkmark}         &  \multicolumn{1}{c|}{{\checkmark}}         & \multicolumn{1}{c|}{}  & & \multicolumn{1}{c|}{\textbf{71.4}}  & \multicolumn{1}{c|}{\textbf{33.4}}  & \multicolumn{1}{c|}{\textbf{60.4}} & \multicolumn{1}{c|}{\underline{82.5}}  & \multicolumn{1}{c|}{63.6}  & \multicolumn{1}{c|}{\underline{81.0}}        \\ 
 \cline{1-13} 

\multicolumn{1}{|c|}{}                 & \multicolumn{1}{c|}{}                           &           & \multicolumn{1}{c|}{\checkmark}         &  \multicolumn{1}{c|}{{\checkmark}}         & \multicolumn{1}{c|}{\checkmark}  & & \multicolumn{1}{c|}{66.5}  & \multicolumn{1}{c|}{28.0}  & \multicolumn{1}{c|}{57.9} & \multicolumn{1}{c|}{\textbf{83.5}}  & \multicolumn{1}{c|}{\textbf{69.2}}  & \multicolumn{1}{c|}{\textbf{83.0}}        \\ 
 \cline{1-13}

\multicolumn{1}{|c|}{}                 & \multicolumn{1}{c|}{}                           &    \checkmark       & \multicolumn{1}{c|}{\checkmark}         &  \multicolumn{1}{c|}{{\checkmark}}         & \multicolumn{1}{c|}{}  & & \multicolumn{1}{c|}{66.4}  & \multicolumn{1}{c|}{26.9}  & \multicolumn{1}{c|}{56.1} & \multicolumn{1}{c|}{82.5}  & \multicolumn{1}{c|}{64.8}  & \multicolumn{1}{c|}{80.8}        \\ 
 \cline{1-13} 

\multicolumn{1}{|c|}{}                 & \multicolumn{1}{c|}{}                           &    \checkmark       & \multicolumn{1}{c|}{\checkmark}         &  \multicolumn{1}{c|}{{\checkmark}}         & \multicolumn{1}{c|}{\checkmark}  & & \multicolumn{1}{c|}{67.6}  & \multicolumn{1}{c|}{28.3}  & \multicolumn{1}{c|}{\underline{58.5}} & \multicolumn{1}{c|}{78.0}  & \multicolumn{1}{c|}{\underline{67.0}}  & \multicolumn{1}{c|}{80.2}        \\ 
 \cline{1-13} 

\multicolumn{1}{|c|}{{\checkmark}}                 & \multicolumn{1}{c|}{{\checkmark}}                           &     {\checkmark}      & \multicolumn{1}{c|}{{\checkmark}}         &  \multicolumn{1}{c|}{{\checkmark}}         & \multicolumn{1}{c|}{{\checkmark}}  & {\checkmark} & \multicolumn{1}{c|}{65.7}  & \multicolumn{1}{c|}{28.9}  & \multicolumn{1}{c|}{56.9} & \multicolumn{1}{c|}{79.0}  & \multicolumn{1}{c|}{63.8}  & \multicolumn{1}{c|}{80.5}        \\

\hline
\end{tabular}
\end{center}
\vspace{-3.0mm}
\end{table}

\begin{table}[!t]
\begin{center}

\caption{\textbf{Ablation Study of the control steps of Instance Shader in the sampling process, in training and inference.}}\label{tab:instanceshader_sample_ablation}\vspace{-2mm}
\renewcommand\arraystretch{1.2}
\setlength{\tabcolsep}{8.7pt}
\begin{tabular}{|c||ccc|ccc|}
\hline \thickhline
\rowcolor[HTML]{F5F5F5}                      \multicolumn{1}{|c||}{\cellcolor[HTML]{F5F5F5}Control}
                   & \multicolumn{3}{c|}{\cellcolor[HTML]{F5F5F5}COCO-MIG-BOX}  
                   & \multicolumn{3}{c|}{\cellcolor[HTML]{F5F5F5}COCO-POSITION} 
\\ 
\cline{2-7}  
\cline{2-7}  
\cline{2-7}  
\cline{2-7}  

\rowcolor[HTML]{F5F5F5} 
 \multicolumn{1}{|c||}{\cellcolor[HTML]{F5F5F5}Steps}& \multicolumn{1}{c|}{\cellcolor[HTML]{F5F5F5}ISR} & \multicolumn{1}{c|}{\cellcolor[HTML]{F5F5F5}SR}  & \multicolumn{1}{c|}{\cellcolor[HTML]{F5F5F5}MIoU}  & \multicolumn{1}{c|}{\cellcolor[HTML]{F5F5F5}SR} & \multicolumn{1}{c|}{\cellcolor[HTML]{F5F5F5}AP} & \multicolumn{1}{c|}{\cellcolor[HTML]{F5F5F5}MIoU}                   \\ 
 \hline\hline

\multicolumn{1}{|c||}{30\%}         & \multicolumn{1}{c|}{66.9}  & \multicolumn{1}{c|}{29.1} & \multicolumn{1}{c|}{55.1}  & \multicolumn{1}{c|}{76.8}  & \multicolumn{1}{c|}{52.3}   & \multicolumn{1}{c|}{75.8} 
\\
 \cline{1-7} 

 \multicolumn{1}{|c||}{40\%}         & \multicolumn{1}{c|}{\underline{70.9}}  & \multicolumn{1}{c|}{\underline{33.1}} & \multicolumn{1}{c|}{\underline{59.4}}  & \multicolumn{1}{c|}{80.9}  & \multicolumn{1}{c|}{60.9}   & \multicolumn{1}{c|}{79.5} 
\\
 \cline{1-7} 

 \multicolumn{1}{|c||}{50\%}         & \multicolumn{1}{c|}{\textbf{71.4}}  & \multicolumn{1}{c|}{\textbf{33.4}} & \multicolumn{1}{c|}{\textbf{60.4}}  & \multicolumn{1}{c|}{\textbf{82.5}}  & \multicolumn{1}{c|}{\underline{63.6}}   & \multicolumn{1}{c|}{\textbf{81.0}}  
\\
 \cline{1-7}

 \multicolumn{1}{|c||}{60\%}         & \multicolumn{1}{c|}{69.8}  & \multicolumn{1}{c|}{31.2} & \multicolumn{1}{c|}{59.1}  & \multicolumn{1}{c|}{\underline{81.3}}  & \multicolumn{1}{c|}{\textbf{63.7}}   & \multicolumn{1}{c|}{\underline{80.6}}  
\\
 \cline{1-7}

\hline
\end{tabular}
\end{center}
\vspace{-3.0mm}
\end{table}

\vspace{-3.0mm}
\subsection{Compare with State-of-the-art Competitors}\label{sec:compare}

\noindent \textbf{COCO-MIG-BOX}. As shown in Tab.~\ref{tab:coco_mig_box}, 
compared to the previous SOTA method, RECO~\cite{reco}, MIGC surpasses it by \textbf{12.3}\% in Instance Success Ratio and \textbf{8.9} in MIoU. 
With the introduction of the Refined Shader, MIGC++ can achieve more precise position-and-attribute control, which enhances the performance of MIGC by \textbf{2.2}\% in Instance Success Ratio and \textbf{3.9}\% in MIoU. 
Additionally, the table shows that as the number of instances increases, the Image Success Ratio of previous SOTA methods is low at L6, for example, GLIGEN~\cite{gligen} is \textbf{0}\%, InstanceDiffusion~\cite{instancediffusion} is \textbf{3.6}\%, and RECO is \textbf{11.2}\%, although MIGC++ raises the success ratio to \textbf{21.6}\%, which still leaves significant room for improvement in the MIG task. 
Fig.~\ref{fig:cocomig_box_vis} shows that both MIGC and MIGC++ exhibit precise control over position and attributes, while MIGC++ provides enhanced control of details, such as successfully generating a ``yellow spoon''.

\noindent \textbf{COCO-MIG-MASK}. 
\textblue{As illustrated in Tab.~\ref{tab:coco_mig_mask}, MIGC++ extends the application of MIGC by allowing masks to describe the fine-grained positions of instances, achieving a \textbf{17.2}\% improvement in MIoU. Compared to Instance Diffusion ~\cite{instancediffusion}, MIGC++ has improved by \textbf{16}\% in Instance Success Ratio, \textbf{12.5}\% in Image Success Ratio, and \textbf{13.3}\% in MIoU.} 
Qualitative comparisons in Fig.\ref{fig:cocomig_mask_vis} further elucidate these metrics, where MIGC++ ensures each instance accurately follows the shapes specified by the masks and has the correct color attributes. As a plug-and-play controller, we also demonstrate the effectiveness of MIGC++ on RV6.0 (i.e., a model fine-tuned from SD1.5), which can generate more realistic images.

\begin{table}[!t]
\begin{center}

\caption{\textbf{Ablation Study of the proposed component on the COCO-Position benchmark.}}\label{tab:component_ablation}\vspace{-2mm}
\renewcommand\arraystretch{1.2}
\setlength{\tabcolsep}{4.5pt}
\begin{tabular}{|c|ccc||ccc|ccc|}
\hline \thickhline
\rowcolor[HTML]{F5F5F5}      
\multicolumn{1}{|c|}{} &
\multicolumn{3}{|c||}{\cellcolor[HTML]{F5F5F5}Component}
                   & \multicolumn{3}{c|}{\cellcolor[HTML]{F5F5F5}MIGC}  
                   & \multicolumn{3}{c|}{\cellcolor[HTML]{F5F5F5}MIGC++} 
\\ 
\cline{2-10} 

\rowcolor[HTML]{F5F5F5} 
\multicolumn{1}{|c|}{\multirow{-2}{*}{num}} &
 \multicolumn{1}{|c|}{\cellcolor[HTML]{F5F5F5}SAC}& \multicolumn{1}{|c|}{\cellcolor[HTML]{F5F5F5}EA}& \multicolumn{1}{|c||}{\cellcolor[HTML]{F5F5F5}LA}& \multicolumn{1}{c|}{\cellcolor[HTML]{F5F5F5}SR} & \multicolumn{1}{c|}{\cellcolor[HTML]{F5F5F5}AP}  & \multicolumn{1}{c|}{\cellcolor[HTML]{F5F5F5}MIoU}  & \multicolumn{1}{c|}{\cellcolor[HTML]{F5F5F5}SR} & \multicolumn{1}{c|}{\cellcolor[HTML]{F5F5F5}AP} & \multicolumn{1}{c|}{\cellcolor[HTML]{F5F5F5}MIoU}                   \\ 
 \hline\hline

\multicolumn{1}{|c|}{\ding{172}} &
\multicolumn{1}{|c|}{}         &  \multicolumn{1}{|c|}{}         &\multicolumn{1}{|c||}{}         &\multicolumn{1}{c|}{7.7}  & \multicolumn{1}{c|}{0.9} & \multicolumn{1}{c|}{22.7}  & \multicolumn{1}{c|}{7.7}  & \multicolumn{1}{c|}{1.4}   & \multicolumn{1}{c|}{23.4} 
\\
 \cline{1-10} 

 \multicolumn{1}{|c|}{\ding{173}} &
\multicolumn{1}{|c|}{\checkmark}         &  \multicolumn{1}{|c|}{}         &\multicolumn{1}{|c||}{}         &\multicolumn{1}{c|}{12.1}  & \multicolumn{1}{c|}{1.9} & \multicolumn{1}{c|}{29.6}  & \multicolumn{1}{c|}{12.4}  & \multicolumn{1}{c|}{2.5}   & \multicolumn{1}{c|}{31.2} 
\\
 \cline{1-10} 

  \multicolumn{1}{|c|}{\ding{174}} &
\multicolumn{1}{|c|}{\checkmark}         &  \multicolumn{1}{|c|}{}         &\multicolumn{1}{|c||}{\checkmark}         &\multicolumn{1}{c|}{34.7}  & \multicolumn{1}{c|}{11.0} & \multicolumn{1}{c|}{44.1}  & \multicolumn{1}{c|}{35.5}  & \multicolumn{1}{c|}{12.9}   & \multicolumn{1}{c|}{45.7} 
\\
 \cline{1-10} 

  \multicolumn{1}{|c|}{\ding{175}} &
\multicolumn{1}{|c|}{\checkmark}         &  \multicolumn{1}{|c|}{\checkmark}         &\multicolumn{1}{|c||}{}         &\multicolumn{1}{c|}{\underline{80.2}}  & \multicolumn{1}{c|}{\underline{53.0}} & \multicolumn{1}{c|}{\underline{76.6}}  & \multicolumn{1}{c|}{\underline{82.3}}  & \multicolumn{1}{c|}{57.4}   & \multicolumn{1}{c|}{\underline{79.3}} 
\\
 \cline{1-10} 

   \multicolumn{1}{|c|}{\ding{176}} &
\multicolumn{1}{|c|}{}         &  \multicolumn{1}{|c|}{\checkmark}         &\multicolumn{1}{|c||}{\checkmark}         &\multicolumn{1}{c|}{78.1}  & \multicolumn{1}{c|}{52.1} & \multicolumn{1}{c|}{75.5}  & \multicolumn{1}{c|}{80.9}  & \multicolumn{1}{c|}{\underline{59.2}}   & \multicolumn{1}{c|}{79.2} 
\\
 \cline{1-10} 

    \multicolumn{1}{|c|}{\ding{177}} &
\multicolumn{1}{|c|}{\checkmark}         &  \multicolumn{1}{|c|}{\checkmark}         &\multicolumn{1}{|c||}{\checkmark}         &\multicolumn{1}{c|}{\textbf{80.3}}  & \multicolumn{1}{c|}{\textbf{54.7}} & \multicolumn{1}{c|}{\textbf{77.4}}  & \multicolumn{1}{c|}{\textbf{82.5}}  & \multicolumn{1}{c|}{\textbf{63.6}}   & \multicolumn{1}{c|}{\textbf{81.0}} 
\\
 \cline{1-10}

\hline
\end{tabular}
\end{center}
\vspace{-3.5mm}
\end{table}
\begin{figure}[t]
    \centering
    \includegraphics[width=1.0\linewidth]{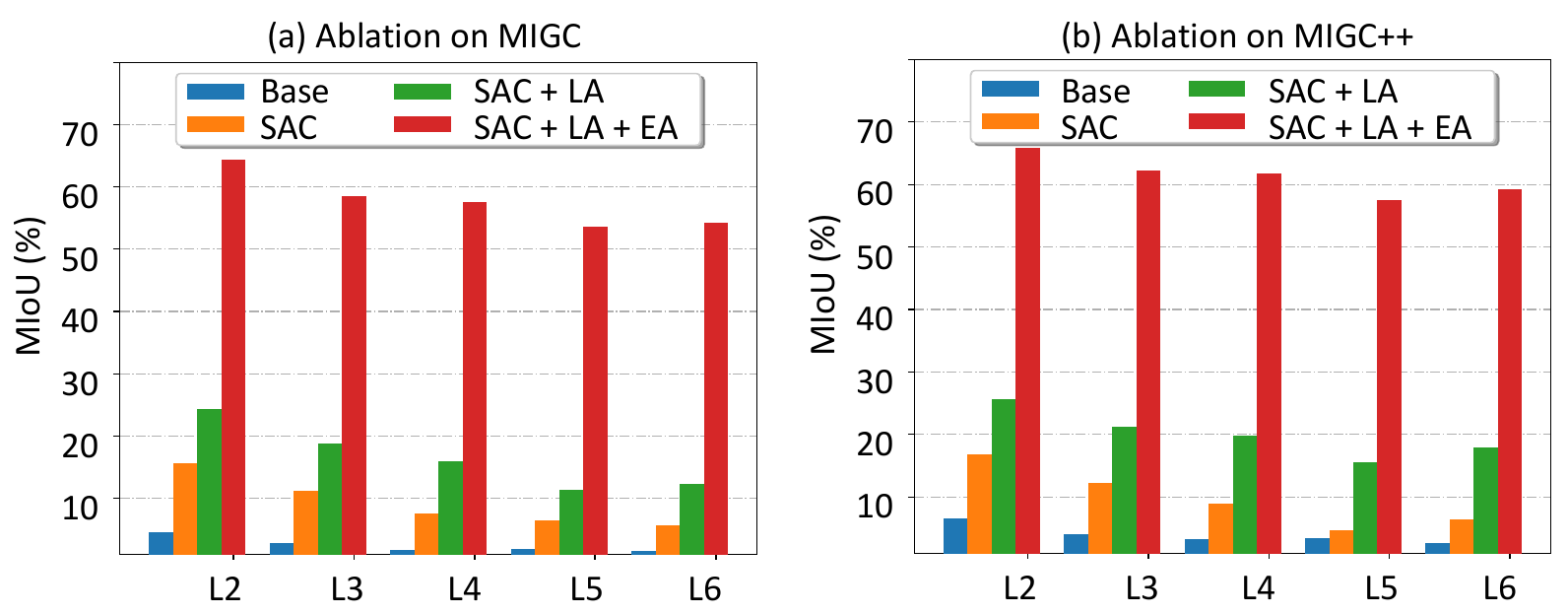}

\vspace{-3.5mm}
    \caption{\textbf{Ablation Study of the proposed component on the COCO-MIG benchmark (\S\ref{sec:ablation}).}}
    \label{fig:cocomig_component_ablation}
    
\vspace{-3.5mm}

\end{figure}

\noindent\textblue{ \textbf{Robustness to Increasing Instance Number}. 
Results from Tab.~\ref{tab:coco_mig_box} and Tab.~\ref{tab:coco_mig_mask} reveal that MIGC++ maintains more consistent performance compared to previous methods as the number of instances increases, thanks to two key enhancements: 1) The divide-and-conquer framework allows MIGC++ to break down multi-instance generation into independent subtasks, which minimizes the impact on every single instance as the number of instances grows—e.g., while InstanceDiffusion saw a \textbf{16.5}\% ISR drop from L2 to L6, MIGC++ experienced only a \textbf{6.2}\% ISR drop in Tab.~\ref{tab:coco_mig_mask}. 2) The introduction of a refined shader, replacing the remaining Cross-Attention layers, secures independent shading of each instance throughout the network’s forward process—where MIGC's performance decreased by \textbf{8.5}\% from L2 to L6, MIGC++ saw a drop of just \textbf{3.3}\% in Tab.~\ref{tab:coco_mig_box}.
}

\noindent \textbf{COCO-POSITION}.
Tab.~\ref{tab:coco_pos} shows that MIGC and MIGC++ achieve results close to those of Real Image and exceed all other methods in positional accuracy. Compared to GLIGEN~\cite{gligen}, MIGC has improved by \textbf{9.77}\% in Success Ratio (SR), \textbf{5.77} in Mean Intersection over Union (MIoU), and \textbf{14.01} in Average Precision (AP). With the introduction of the Refined Shader, MIGC++ further enhances these metrics based on MIGC, with an increase of \textbf{2.25}\% in SR, \textbf{3.64} in MIoU, and \textbf{8.87} in AP. MIGC and MIGC++ also achieve similar FID scores compared to the SD~\cite{stablediffusion}, highlighting that MIGC can enhance position control capabilities without compromising image quality. Fig.~\ref{fig:coco_pos_vis} illustrates that MIGC and MIGC++ display superior control over instance position, consistently placing items accurately within designated position. Additionally, MIGC++ offers enhanced detail in instance generation compared to MIGC, notably in the more refined depiction of the apple and orange.

\noindent \textbf{DrawBench}.
Tab.~\ref{tab:DrawBench} presents that text-to-image methods achieve only a \textbf{23.1}\% accuracy in position control and \textbf{30.9}\% in count control. 
By utilizing GPT-4~\cite{layoutgpt,gpt4}, which analyzes image descriptions and generates instance descriptions, improvements can be achieved. 
For example, GLIGEN's accuracy in position control increased from \textbf{23.1}\% to \textbf{78.8}\%. However, these methods are prone to instance attribute leakage, achieving only \textbf{65.5}\% accuracy in attribute control. Building on this, our proposed model, MIGC++, effectively avoids the issue of attribute leakage, enhancing attribute control accuracy from \textbf{65.5}\% to \textbf{98.5}\%. Additionally, it improves position control by \textbf{14.3}\% and count control by \textbf{15.8}\% compared to the previous SOTA method.

\noindent \textbf{Multimodal-MIG}. 
Fig.~\ref{fig:multimodal_mig_vis} compares the results of MIGC++ with tuning-free methods ELITE\cite{elite} and SSR-Encoder~\cite{ssr}, as well as the tuning-required Custom Diffusion~\cite{customdiffusion}, on the Multimodal-MIG benchmark. MIGC++ preserves the ID of the reference image and accurately generates other instances as described in the text. For instance, MIGC++ uniquely succeeded in generating a black apple and a real duck in the first and second rows, while accurately positioning them as specified. In the third row, MIGC++ also excelled in controlling the number of instances along with maintaining ID accuracy, a feat not matched by other methods. Furthermore, a user study involving a pairwise comparison of the tuning-free methods showed that MIGC++ achieved superior text-image alignment and maintained better similarity to the reference image, as detailed in Fig.~\ref{fig:multimodal_mig_user}.

\vspace{-3.0mm}
\subsection{Ablation Study}\label{sec:ablation}

\noindent \textbf{Deployment Position of Instance Shader}.
Tab.~\ref{tab:instanceshader_pos_ablation} shows that the best control over attributes and positions is achieved when the Instance Shader is deployed in the mid and up-1 blocks of U-net, with up-1 being particularly crucial. Adding the shader to up-2 enhances position control but might compromise attribute management. Conversely, including it in down-2 does not improve and may even reduce control effectiveness. Deploying the shader across all blocks reduces its effectiveness in control, likely due to interference from detailed feature processing in the shallow blocks.

\noindent \textbf{Control steps of Instance Shader}.
Tab.~\ref{tab:instanceshader_sample_ablation} shows that activating the Instance Shader during the first 50\% of sampling steps optimally controls position and attributes, as these early stages are critical for establishing the image's layout and attributes. Extending activation to 60\% in training can undermine control during the crucial initial 50\%, resulting in diminished capabilities. Similarly, limiting control to just 30\% or 40\% of the steps also leads to reduced effectiveness.

\begin{figure}[t]
    \centering
    \includegraphics[width=1.0\linewidth]{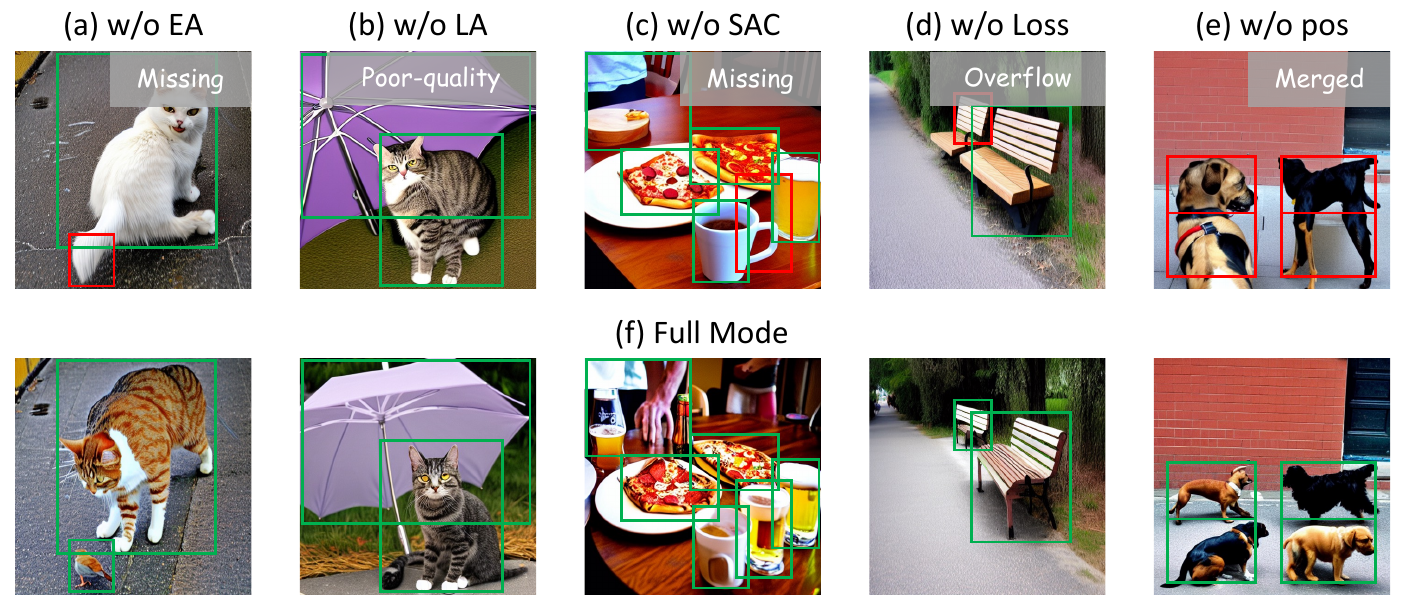}

\vspace{-3.5mm}
    \caption{\textbf{Qualitative Outcomes from Ablation Studies on Proposed Components. (\S\ref{sec:ablation})} "pos" means the position embedding used in the EA mechanism (\S\ref{sec:ea}, \S\ref{sec:mmea}). We mark incorrectly generated instances with \textcolor{red}{red boxes} and correctly with \textcolor{green}{green boxes}.}
    \label{fig:ablation_vis}
    
\vspace{-3.5mm}

\end{figure}

\begin{table}[!t]
\begin{center}

\caption{\textbf{Quantitative Analysis of Diverse Deployments for the Refined Shader (\S\ref{sec:ablation}).}}\label{tab:refinedshader_pos_ablation}\vspace{-2mm}
\renewcommand\arraystretch{1.2}
\setlength{\tabcolsep}{3pt}
\begin{tabular}{|ccc|c|ccc||ccc|ccc|}
\hline \thickhline
\rowcolor[HTML]{F5F5F5}                      \multicolumn{3}{|c|}{\cellcolor[HTML]{F5F5F5}down}      & \multicolumn{1}{c|}{\cellcolor[HTML]{F5F5F5}mid}                                        & \multicolumn{3}{c||}{\cellcolor[HTML]{F5F5F5}up}
                   & \multicolumn{3}{c|}{\cellcolor[HTML]{F5F5F5}COCO-MIG-BOX}  
                   & \multicolumn{3}{c|}{\cellcolor[HTML]{F5F5F5}COCO-POSITION} 
\\ 
\cline{1-13} 
\thickhline

\rowcolor[HTML]{F5F5F5} 
 \multicolumn{1}{|c|}{\cellcolor[HTML]{F5F5F5}0} & \multicolumn{1}{c|}{\cellcolor[HTML]{F5F5F5}1} & 2  & \multicolumn{1}{c|}{\cellcolor[HTML]{F5F5F5}0}      & \multicolumn{1}{c|}{\cellcolor[HTML]{F5F5F5}1} & \multicolumn{1}{c|}{\cellcolor[HTML]{F5F5F5}2} & \multicolumn{1}{c||}{\cellcolor[HTML]{F5F5F5}3} & \multicolumn{1}{c|}{\cellcolor[HTML]{F5F5F5}ISR} & \multicolumn{1}{c|}{\cellcolor[HTML]{F5F5F5}SR}  & \multicolumn{1}{c|}{\cellcolor[HTML]{F5F5F5}MIoU}  & \multicolumn{1}{c|}{\cellcolor[HTML]{F5F5F5}SR} & \multicolumn{1}{c|}{\cellcolor[HTML]{F5F5F5}AP} & \multicolumn{1}{c|}{\cellcolor[HTML]{F5F5F5}MIoU}                   \\ 
  \hline\hline

\multicolumn{1}{|c|}{}                 & \multicolumn{1}{c|}{}                           &           & \multicolumn{1}{c|}{{}}         &  \multicolumn{1}{c|}{}         & \multicolumn{1}{c|}{}  & & \multicolumn{1}{c|}{67.2}  & \multicolumn{1}{c|}{28.3}  & \multicolumn{1}{c|}{56.8} & \multicolumn{1}{c|}{79.9}  & \multicolumn{1}{c|}{59.2}  & \multicolumn{1}{c|}{79.4} 
\\
 \cline{1-13}

\multicolumn{1}{|c|}{}                 & \multicolumn{1}{c|}{}                           &           & \multicolumn{1}{c|}{}         &  \multicolumn{1}{c|}{{\checkmark}}         & \multicolumn{1}{c|}{\checkmark}  & \checkmark & \multicolumn{1}{c|}{70.2}  & \multicolumn{1}{c|}{32.0}  & \multicolumn{1}{c|}{59.4} & \multicolumn{1}{c|}{80.5}  & \multicolumn{1}{c|}{60.3}  & \multicolumn{1}{c|}{79.9}        \\ 
 \cline{1-13} 

\multicolumn{1}{|c|}{\checkmark}                 & \multicolumn{1}{c|}{\checkmark}                           &      \checkmark     & \multicolumn{1}{c|}{}         &  \multicolumn{1}{c|}{{}}         & \multicolumn{1}{c|}{}  & & \multicolumn{1}{c|}{69.5}  & \multicolumn{1}{c|}{\underline{32.4}}  & \multicolumn{1}{c|}{58.7} & \multicolumn{1}{c|}{81.4}  & \multicolumn{1}{c|}{\underline{63.1}}  & \multicolumn{1}{c|}{\underline{80.6}}        \\ 
 \cline{1-13} 

\multicolumn{1}{|c|}{}                 & \multicolumn{1}{c|}{}                           &           & \multicolumn{1}{c|}{\checkmark}         &  \multicolumn{1}{c|}{{}}         & \multicolumn{1}{c|}{}  & & \multicolumn{1}{c|}{67.4}  & \multicolumn{1}{c|}{28.3}  & \multicolumn{1}{c|}{56.9} & \multicolumn{1}{c|}{81.0}  & \multicolumn{1}{c|}{60.8}  & \multicolumn{1}{c|}{79.7}        \\ 
 \cline{1-13}

\multicolumn{1}{|c|}{}                 & \multicolumn{1}{c|}{}                           &           & \multicolumn{1}{c|}{\checkmark}         &  \multicolumn{1}{c|}{{\checkmark}}         & \multicolumn{1}{c|}{\checkmark}  & \checkmark & \multicolumn{1}{c|}{\underline{70.4}}  & \multicolumn{1}{c|}{32.3}  & \multicolumn{1}{c|}{\underline{59.5}} & \multicolumn{1}{c|}{81.4}  & \multicolumn{1}{c|}{61.2}  & \multicolumn{1}{c|}{80.1}        \\ 
 \cline{1-13} 

\multicolumn{1}{|c|}{\checkmark}                 & \multicolumn{1}{c|}{\checkmark}                           &    \checkmark       & \multicolumn{1}{c|}{\checkmark}         &  \multicolumn{1}{c|}{{}}         & \multicolumn{1}{c|}{}  & & \multicolumn{1}{c|}{69.9}  & \multicolumn{1}{c|}{32.3}  & \multicolumn{1}{c|}{59.0} & \multicolumn{1}{c|}{\underline{81.6}}  & \multicolumn{1}{c|}{62.7}  & \multicolumn{1}{c|}{\underline{80.6}}        \\ 
 \cline{1-13} 

\multicolumn{1}{|c|}{{\checkmark}}                 & \multicolumn{1}{c|}{{\checkmark}}                           &     {\checkmark}      & \multicolumn{1}{c|}{{\checkmark}}         &  \multicolumn{1}{c|}{{\checkmark}}         & \multicolumn{1}{c|}{{\checkmark}}  & {\checkmark} & \multicolumn{1}{c|}{\textbf{71.4}}  & \multicolumn{1}{c|}{\textbf{33.4}}  & \multicolumn{1}{c|}{\textbf{60.4}} & \multicolumn{1}{c|}{\textbf{82.5}}  & \multicolumn{1}{c|}{\textbf{63.6}}  & \multicolumn{1}{c|}{\textbf{81.0}}        \\

\hline
\end{tabular}
\end{center}
\vspace{-3.0mm}
\end{table}

\noindent \textbf{Enhancement Attention}. 
The comparison between results \ding{173} and \ding{175} highlights that Enhance Attention (EA) is crucial for successful single-instance shading, markedly boosting image generation success rates. This effect is corroborated by the COCO-MIG benchmark data shown in Fig.~\ref{fig:cocomig_component_ablation}. Specifically, Fig.\ref{fig:ablation_vis}(e) demonstrates that omitting position embedding during EA application can lead to instance merging, as exemplified by the merging of two dog-containing boxes into one. Fig.~\ref{fig:ablation_vis}(a) reveals that without EA, a bird vanishes, whereas with EA, the bird is clearly depicted in Fig.~\ref{fig:ablation_vis}(f).

\noindent \textbf{Layout Attention}.
The comparison between Fig. ~\ref{fig:ablation_vis}(b) and Fig.~\ref{fig:ablation_vis}(f) illustrates that Layout Attention (LA) is crucial in bridging shading instances, which enhances the quality and cohesion of the generated images.
Additionally, by comparing results \ding{173} and \ding{174} in Tab.~\ref{tab:component_ablation}, we can observe that LA also plays a role in successfully shading instances, as also demonstrated in Fig.~\ref{fig:cocomig_component_ablation}.

\noindent \textbf{Shading Aggregation Controller}. 
Comparing Fig.~\ref{fig:ablation_vis}(c) and Fig.~\ref{fig:ablation_vis}(f), it can be seen that the Shading Aggregation Controller (SAC) can effectively improve the accuracy of multi-instance shading in complex layout, such as when there is overlap between multiple instances. Comparing results \ding{176} and \ding{177} in Tab.~\ref{tab:component_ablation}, the addition of the SAC further enhances the accuracy of multi-instance shading.

\begin{figure}[tb]
    \centering
    \includegraphics[width=1.0\linewidth]{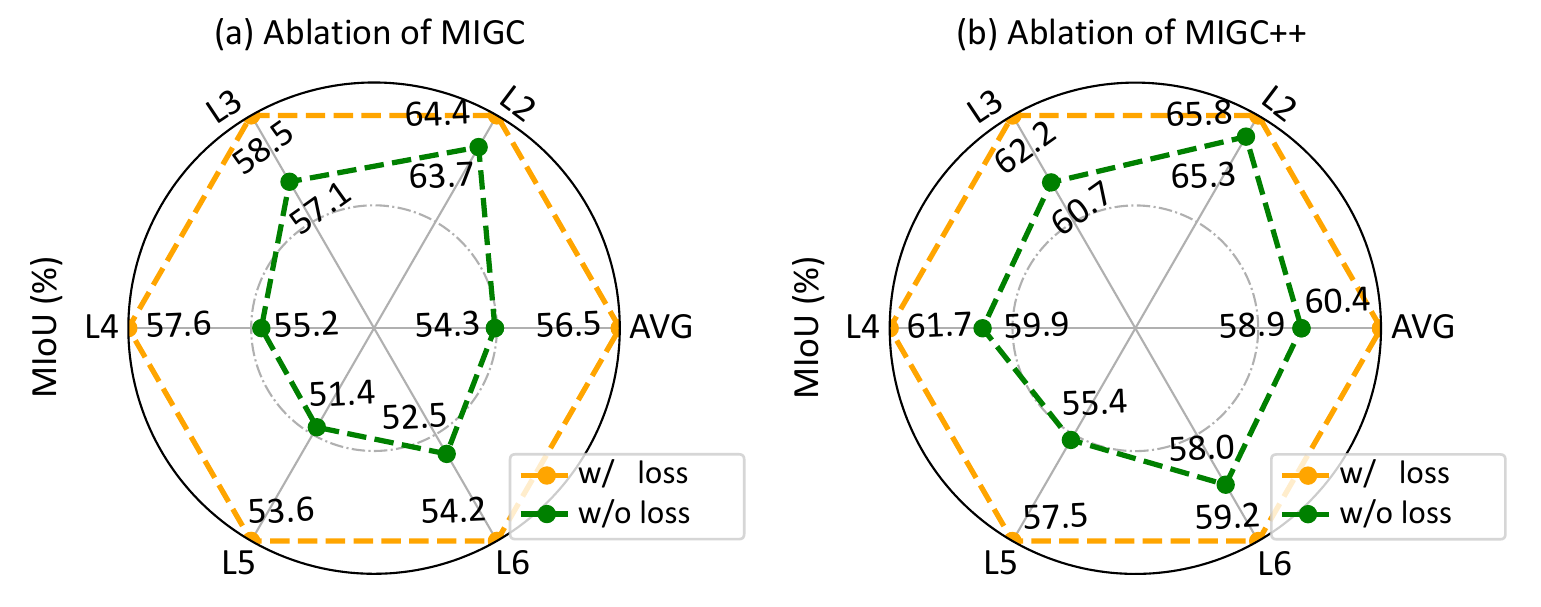}

\vspace{-3.5mm}
    \caption{\textbf{Ablation study of the Inhibition Loss on the COCO-MIG benchmark (\S\ref{sec:ablation}).} 
    \textblue{We investigate the improvements brought by inhibition loss across different instance counts using the default loss weight (see Tab.~\ref{tab:inhb_loss_ablation}).}}
    \label{fig:cocomig_loss_abltion}
    
\vspace{-3.5mm}

\end{figure}

\noindent \textbf{Inhibition Loss}.
Tab.~\ref{tab:inhb_loss_ablation} indicates that using an inhibition loss with a weight of 0.1, as compared to not using this loss, allows MIGC to increase its AP by \textbf{2.23} and MIGC++ by \textbf{2.08}. 
\textblue{Although further increasing the loss weight offers a slight boost in control ability, it somewhat impacts the quality of generated images. For instance, with a higher loss weight (i.e., 1.0), the FID score worsens by \textbf{2.42} for MIGC and by \textbf{0.74} for MIGC++. Since both control ability and image quality are crucial for generative models, we set the default inhibition loss weight to 0.1 to enhance control without affecting image quality.}
Fig.~\ref{fig:cocomig_loss_abltion}, reveals that the advantages of inhibition loss are more pronounced with higher numbers of generated instances. For instance, with two instances, inhibition loss only improves MIoU from \textbf{65.3} to \textbf{65.8} (a \textbf{0.5} increase), but with five instances, it boosts from \textbf{55.4} to \textbf{57.5} (a \textbf{2.1} increase). Comparing Fig.~\ref{fig:ablation_vis}(d) and Fig.~\ref{fig:ablation_vis}(f), it is apparent that inhibition loss more effectively restricts the instances to their intended locations.
\begin{table}[!t]
\begin{center}

\caption{\textbf{Ablation Study of the proposed inhibition loss on the COCO-Position benchmark.} \textblue{We chose 0.1 as the default loss weight because it enhances control ability without affecting image quality (FID$\downarrow$).}}\label{tab:inhb_loss_ablation}\vspace{-2mm}
\renewcommand\arraystretch{1.2}
\setlength{\tabcolsep}{3.9pt}
\begin{tabular}{|c|cccc|cccc|}
\hline \thickhline
\rowcolor[HTML]{F5F5F5}    
\multicolumn{1}{|c||}{loss}
                   & \multicolumn{4}{c|}{\cellcolor[HTML]{F5F5F5}MIGC}  
                   & \multicolumn{4}{c|}{\cellcolor[HTML]{F5F5F5}MIGC++} 
\\ 
\cline{2-9} 
\cline{2-9} 
\cline{2-9} 
\cline{2-9} 

\rowcolor[HTML]{F5F5F5}  \multicolumn{1}{|c||}{{weight}}& \multicolumn{1}{c|}{\cellcolor[HTML]{F5F5F5}SR} & \multicolumn{1}{c|}{\cellcolor[HTML]{F5F5F5}AP}  & \multicolumn{1}{c|}{\cellcolor[HTML]{F5F5F5}MIoU} & \multicolumn{1}{c|}{\cellcolor[HTML]{F5F5F5}FID}  & \multicolumn{1}{c|}{\cellcolor[HTML]{F5F5F5}SR} & \multicolumn{1}{c|}{\cellcolor[HTML]{F5F5F5}AP} & \multicolumn{1}{c|}{\cellcolor[HTML]{F5F5F5}MIoU} & \multicolumn{1}{c|}{\cellcolor[HTML]{F5F5F5}FID}  
\\ 
 \hline\hline

\multicolumn{1}{|c||}{0.0}         &\multicolumn{1}{c|}{80.20}  & \multicolumn{1}{c|}{52.46} & \multicolumn{1}{c|}{77.03} & \multicolumn{1}{c|}{\underline{24.73}}  & \multicolumn{1}{c|}{82.12}  & \multicolumn{1}{c|}{61.48}   & \multicolumn{1}{c|}{80.45} & \multicolumn{1}{c|}{\underline{}{24.56}}
\\
 \cline{1-9} 

\multicolumn{1}{|c||}{0.1}         &\multicolumn{1}{c|}{\underline{80.29}}  & \multicolumn{1}{c|}{\underline{54.69}} & \multicolumn{1}{c|}{\underline{77.38}} & \multicolumn{1}{c|}{\textbf{24.52}}  & \multicolumn{1}{c|}{\underline{82.54}}  & \multicolumn{1}{c|}{\underline{63.56}}   & \multicolumn{1}{c|}{\underline{81.02}} & \multicolumn{1}{c|}{\textbf{24.42}}
\\
 \cline{1-9} 

\multicolumn{1}{|c||}{1.0}         &\multicolumn{1}{c|}{\textbf{80.61}}  & \multicolumn{1}{c|}{\textbf{55.62}} & \multicolumn{1}{c|}{\textbf{77.79}} & \multicolumn{1}{c|}{26.94}  & \multicolumn{1}{c|}{\textbf{82.73}}  & \multicolumn{1}{c|}{\textbf{64.12}}   & \multicolumn{1}{c|}{\textbf{81.33}} & \multicolumn{1}{c|}{25.16}
\\


\hline
\end{tabular}
\end{center}
\vspace{-3.0mm}
\end{table}
\begin{figure}[t]
    \centering
    \includegraphics[width=1.0\linewidth]{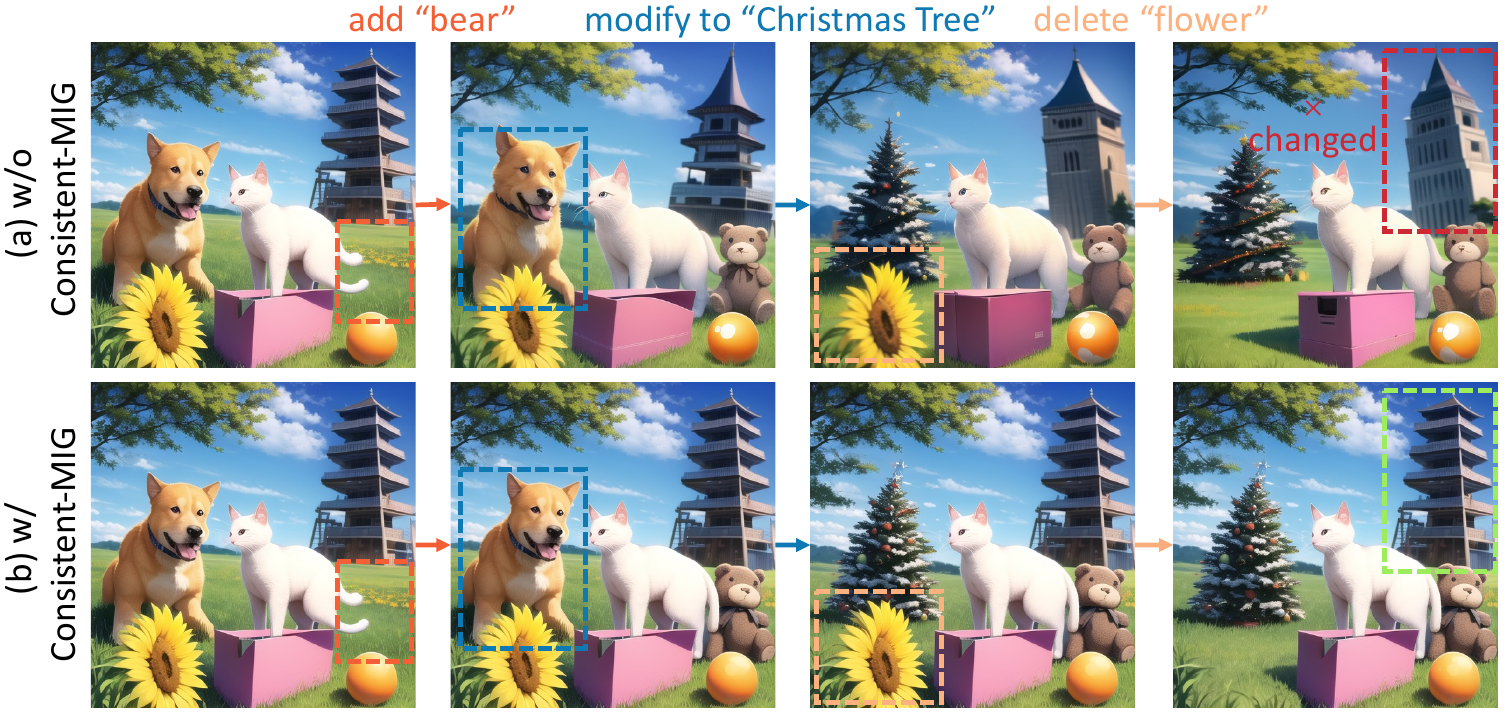}

\vspace{-3.5mm}
    \caption{\textbf{Ablation study of Consistent-MIG (\S\ref{sec:ablation}).}}
    \label{fig:consistent_mig_ablation}
    
\vspace{-3.5mm}

\end{figure}
\begin{figure}[t]
    \centering
    \includegraphics[width=1.0\linewidth]{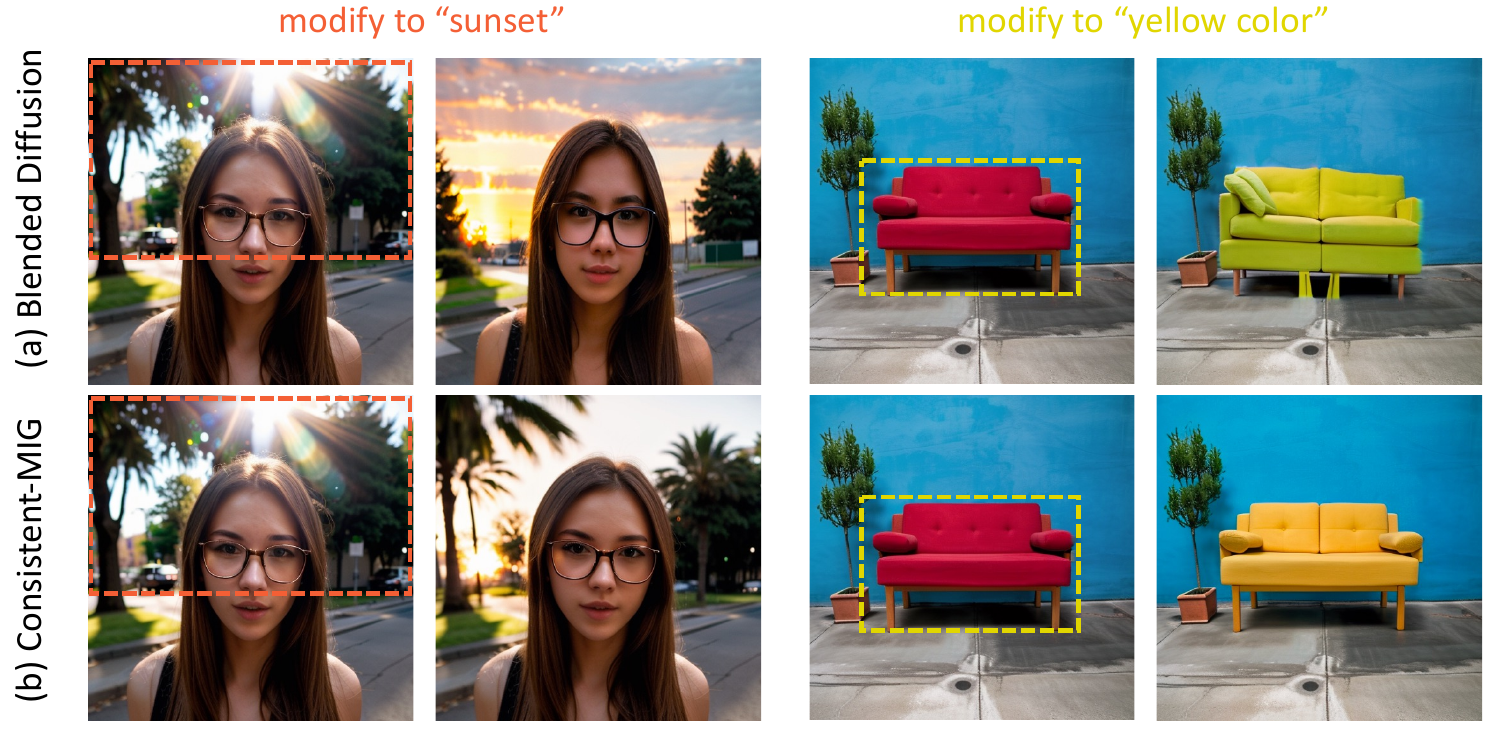}

\vspace{-3.5mm}
    \caption{
\textbf{Comparing Consistent-MIG vs. BLD~\cite{blenddiff}} (\S~\ref{sec:ablation}).}
    \label{fig:consistent_MIG_vs_bld}
    
\vspace{-3.5mm}

\end{figure}
\begin{figure}[t!]
    \centering
    \includegraphics[width=1.0\linewidth]{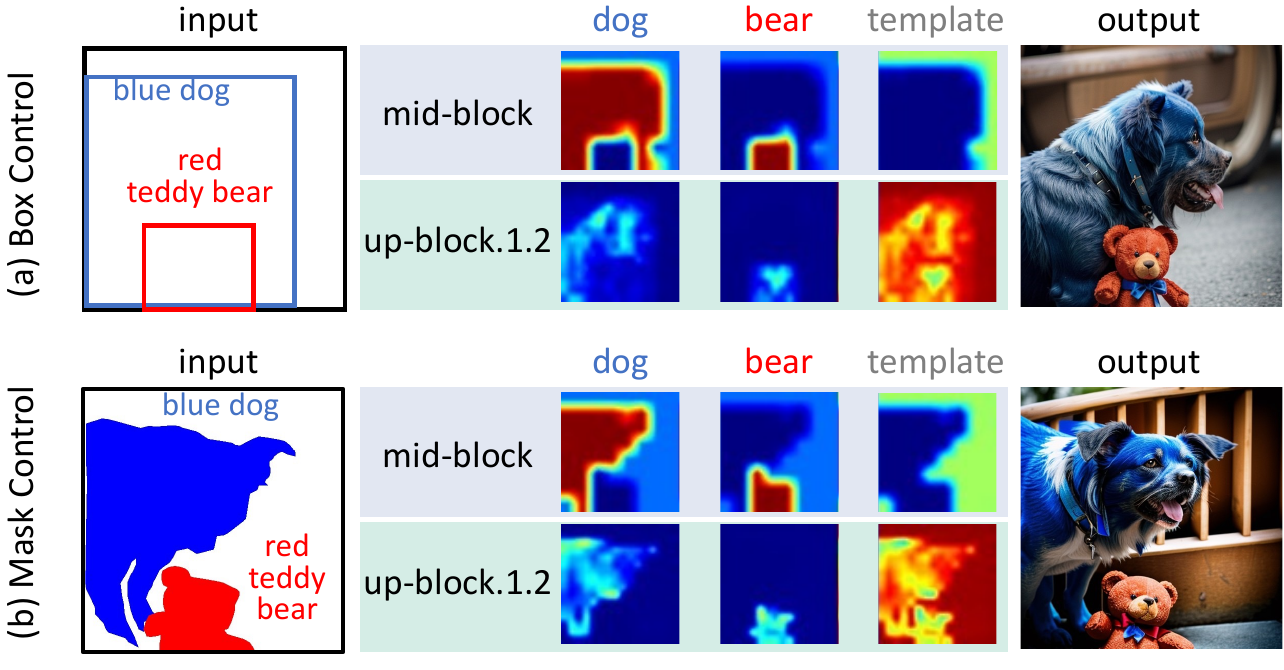}

\vspace{-3.5mm}
    \caption{\textbf{Aggregation weights of SAC (\S\ref{sec:sac}).}}
    \label{fig:vis_aggregation}
    
\vspace{-3.5mm}

\end{figure}

\noindent \textbf{Refined Shader}.
Tab.~\ref{tab:refinedshader_pos_ablation} shows that employing the Refined Shader across all blocks of U-net significantly enhances the Success Ratio, with a \textbf{4.2}\% increase in COCO-MIG and a \textbf{2.6}\% rise in COCO-POSITION. Analysis indicates that the up-blocks are key for attribute control, contributing to notable gains in COCO-MIG, whereas the down-blocks are essential for improving positional accuracy, thus boosting performance in COCO-POSITION. Fig.~\ref{fig:migc_overview} demonstrates that the Refined Shader mitigates attribute confusion among semantically similar instances. Additionally, Fig.~\ref{fig:refinedshader_effect} reveals that instances shaded with the Refined Shader more accurately reflect the details of the reference image.

\noindent \textbf{Consistent-MIG}. Fig.~\ref{fig:consistent_mig_ablation} illustrates that the Consistent-MIG algorithm maintains consistency in unmodified areas across iterations, as seen in Fig.~\ref{fig:consistent_mig_ablation}(b), where results from previous iterations are preserved. Without Consistent-MIG, as shown in Fig.~\ref{fig:consistent_mig_ablation}(a), unaltered areas continue to change, exemplified by the tower in the background altering with each iteration. Fig.~\ref{fig:consistent_MIG_vs_bld} compares Consistent-MIG with Blended Diffusion~\cite{blenddiff}. While Blended Diffusion fails to maintain the identity of people and alters the structure of objects like chairs, Consistent-MIG more effectively preserves the integrity and identity of modified instances.

\vspace{-3.0mm}
\subsection{Visualization of Multi-Instance Shading}\label{sec:vis_multi_shading}
Fig.~\ref{fig:vis_aggregation}(a) and Fig.~\ref{fig:vis_aggregation}(b) display aggregation weights at time t=0.6 (starting at t=1.0). 
In the first block, i.e., the mid-block, significant weights are assigned to instances within their specific areas, while a shading template integrates these instances into the background. Through the sequential stages of mid-block, up-block-1-0, and up-block-1-1, multi-instance shading is effectively executed, with each instance's structure clearly delineated in the aggregation maps by up-block-1-2, and this shader primarily refines the shading template, enhancing overall image cohesion. Fig.~\ref{fig:vis_aggregation}(a) illustrates that overlapping instances trigger adaptive adjustments by the shader, ensuring unique spatial allocations. Conversely, as shown in Fig.~\ref{fig:vis_aggregation}(b), when instances are defined by a mask, the shader allocates weights precisely according to the mask.



\section{Conclusion}
In this study, we introduce the Multi-Instance Generation (MIG) task, addressing key challenges in attribute leakage, restricted instance descriptions, and iterative capabilities. We propose the MIGC method, employing a divide-and-conquer strategy to decompose multi-instance shading into manageable sub-tasks. This approach prevents attribute leakage by merging attribute-correct solutions from these sub-tasks. Enhancing MIGC, MIGC++ allows instance attributes to be defined via text and images, utilizing boxes and masks for precise placement and introducing a Refined Shader for detailed attribute accuracy. To advance iterative MIG performance, we developed the Consistent-MIG algorithm, ensuring consistency in unaltered areas and identity consistency across modified instances. We established the COCO-MIG and Multimodal-MIG benchmarks to evaluate model efficacy. Testing across five benchmarks confirms that MIGC and MIGC++ outperform existing methods, offering precise control over position, attributes, and quantity. We anticipate these methodologies and benchmarks will propel further research on MIG and associated tasks, setting a new standard for future explorations.




%



\ifCLASSOPTIONcaptionsoff
  \newpage
\fi



\small{
\bibliographystyle{IEEEtran}
\bibliography{reference}
}
\vspace{-10mm}

\begin{IEEEbiography} [{\includegraphics[width=0.9in,clip,keepaspectratio]{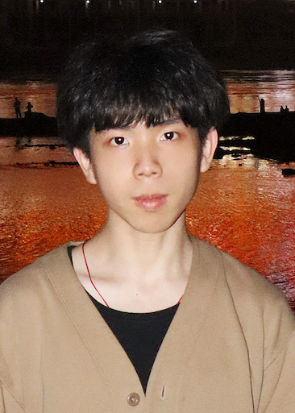}}]{Dewei Zhou} received the B.E. degree in the School of Computer and Artificial Intelligence from Zhengzhou University, Zhengzhou, China, in 2021. He is currently pursuing a Ph.D. degree in the School of Computer Science and Technology, Zhejiang University. His research interests include image generation, diffusion models, and image enhancement.
\end{IEEEbiography}

\vspace{-10mm}

\begin{IEEEbiography} [{\includegraphics[width=0.9in,clip,keepaspectratio]{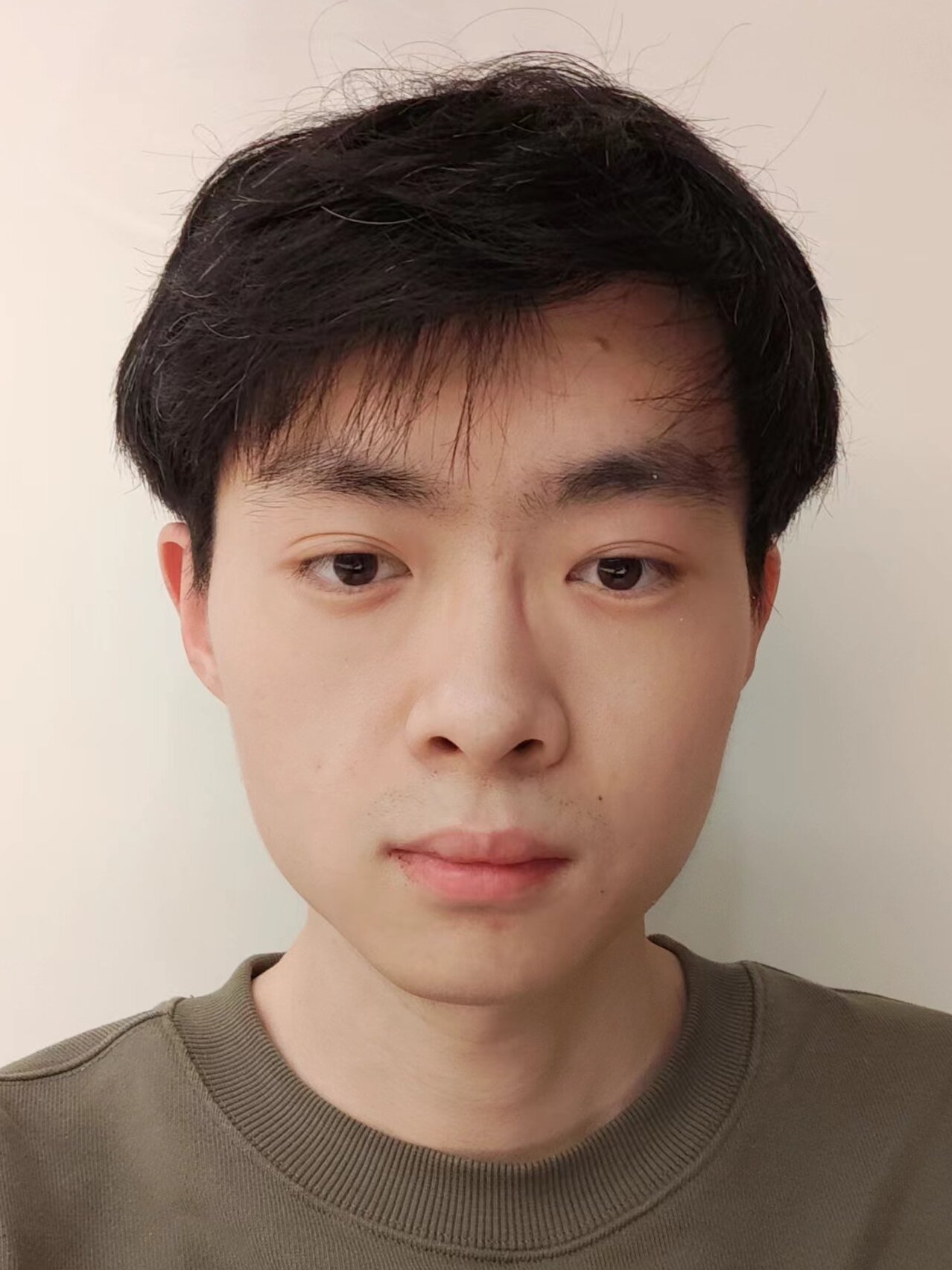}}] {You Li} received the B.E. degree in Computer Science and Technology from Zhejiang University, Hangzhou, China, in 2022. He is currently pursuing a Ph.D. degree in Zhejiang University of China. His research interests include Image generation, diffusion models.\end{IEEEbiography}
\vspace{-10mm}

\begin{IEEEbiography} [{\includegraphics[width=0.9in,clip,keepaspectratio]{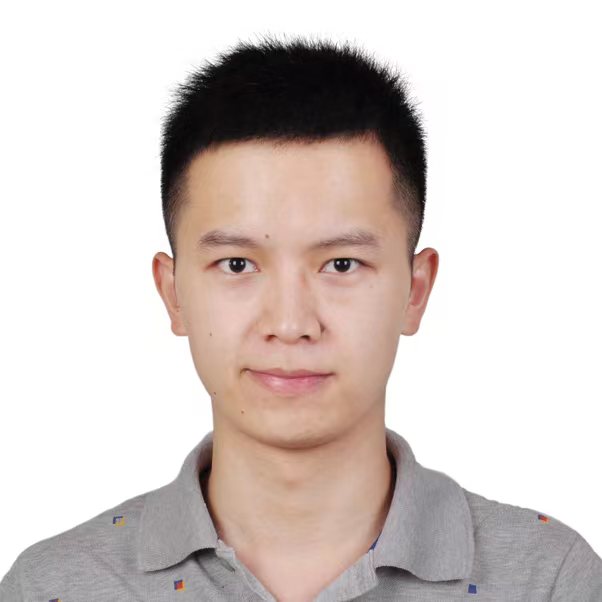}}] {Fan Ma} is currently a research fellow at Zhejiang University. He previously received his Ph.D. degree from the University of Technology Sydney. His research interests include multimodal learning and temporal modeling. His algorithms and methods have significantly impacted model pre-training with limited and imperfect training data.
\end{IEEEbiography}

\vspace{-10mm}

\begin{IEEEbiography} [{\includegraphics[width=0.9in,clip,keepaspectratio]{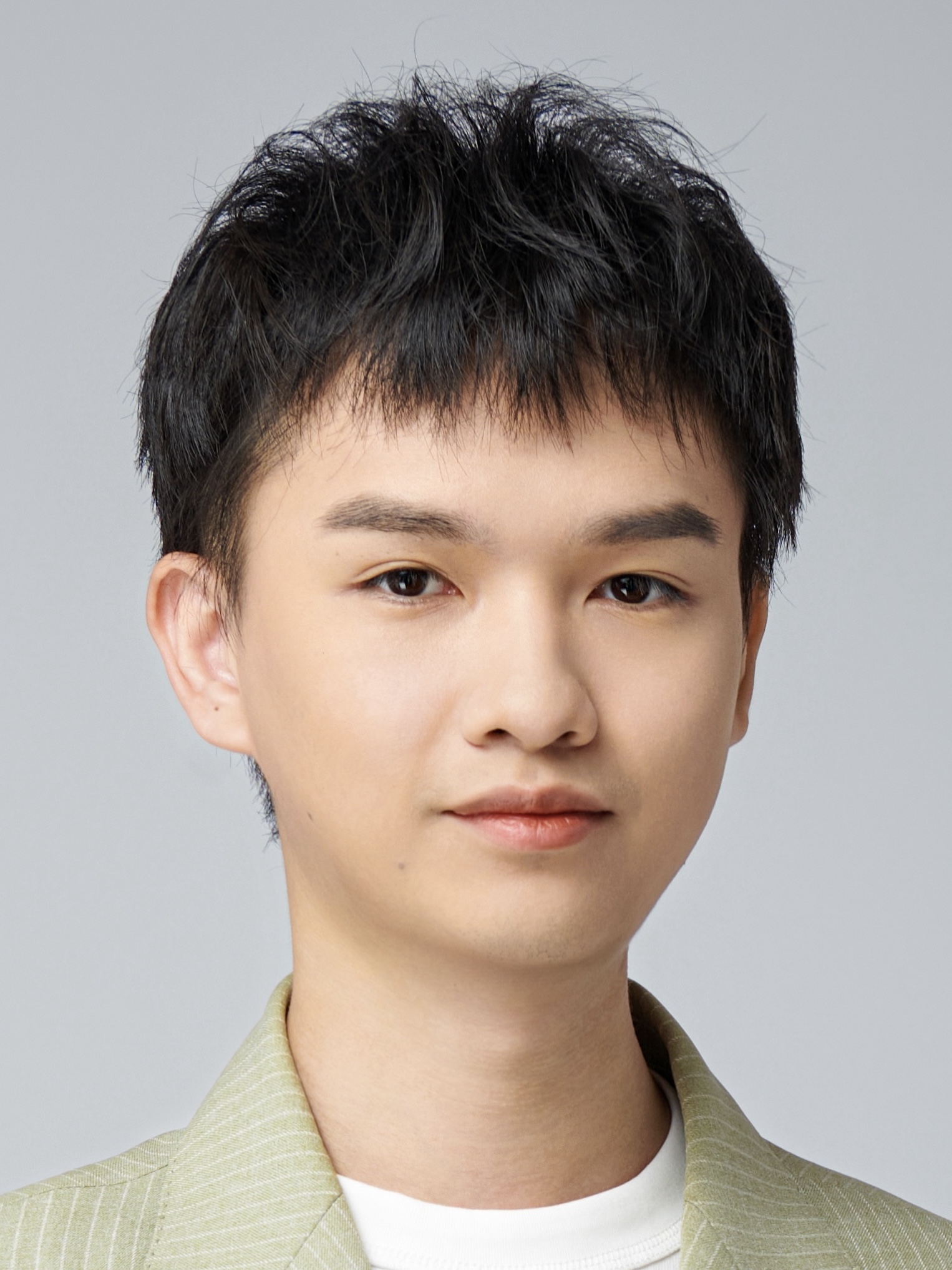}}]{Zongxin Yang} received his Bachelor’s degree in Engineering (BE) from the University of Science and Technology of China in 2018 and earned his Ph.D. in Computer Science from the University of Technology Sydney in 2021. He is currently a postdoctoral researcher at Harvard University. His research interests include multi-modal learning, vision generation, and the intersection of biomedical science and AI.
\end{IEEEbiography}

\vspace{-10mm}

\begin{IEEEbiography} [{\includegraphics[width=0.9in,clip,keepaspectratio]{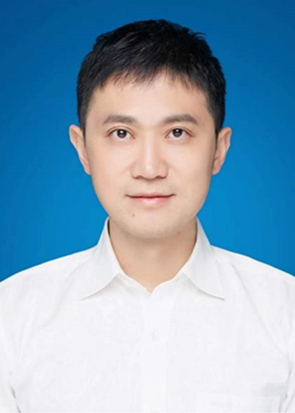}}]{Yi Yang} (Senior Member, IEEE) received the PhD
degree from Zhejiang University, in 2010. He is
a distinguished professor with Zhejiang University,
China. His current research interests include machine
learning and its applications to multimedia content
analysis and computer vision, such as multimedia
retrieval and video content understanding. He received the Australia Research Council Early Career
Researcher Award, the Australia Computing Society,
the Google Faculty Research Award, and the AWS
Machine Learning Research Award Gold Disruptor
Award.
\end{IEEEbiography}
%


%







\newpage

\end{document}